\documentclass[lettersize,journal]{IEEEtran}
\usepackage{amsmath,amsfonts}
\usepackage{algorithmic}
\usepackage{algorithm}
\usepackage{array}
\usepackage{textcomp}
\usepackage{stfloats}
\usepackage{url}
\usepackage{verbatim}
\usepackage{graphicx}
\usepackage{cite}
\usepackage[caption=false,font=footnotesize,labelfont=rm,textfont=rm]{subfig}
\usepackage{bm}
\usepackage{multirow}
\usepackage{enumitem}
\usepackage{colortbl}
\usepackage{booktabs}    
\usepackage{amssymb}
\usepackage[pagebackref=true,breaklinks=true,colorlinks,bookmarks=false]{hyperref}
\usepackage{threeparttable}

\definecolor{mygray}{gray}{.85}

\newcommand{\tabincell}[2]{\begin{tabular}{@{}#1@{}}#2\end{tabular}}

\newcommand{\etal}{\textit{et al}.}
\newcommand{\ie}{\textit{i}.\textit{e}.}

\begin{document}

\title{PFF-Net: Patch Feature Fitting for Point Cloud Normal Estimation}

\author{
Qing Li~\IEEEmembership{Member, IEEE}, Huifang Feng, Kanle Shi, Yue Gao~\IEEEmembership{Senior Member, IEEE}, Yi Fang, Yu-Shen Liu~\IEEEmembership{Member, IEEE}, Zhizhong Han

\IEEEcompsocitemizethanks{
  \IEEEcompsocthanksitem Qing Li is with the School of Computing and Artificial Intelligence, Southwest Jiaotong University, Chengdu, China. E-mail: qingli@swjtu.edu.cn
  \IEEEcompsocthanksitem Huifang Feng is with the School of Computer and Software Engineering, Xihua University, Chengdu, China. E-mail: fhf@xhu.edu.cn
  \IEEEcompsocthanksitem Yue Gao and Yu-Shen Liu are with the School of Software, Tsinghua University, Beijing, China. E-mail: \{gaoyue, liuyushen\}@tsinghua.edu.cn
  \IEEEcompsocthanksitem Kanle Shi is with Kuaishou Technology, Beijing, China. E-mail: shikanle@kuaishou.com
  \IEEEcompsocthanksitem Yi Fang is with the Center for Artificial Intelligence and Robotics, New York University Abu Dhabi, Abu Dhabi, UAE. E-mail: yfang@nyu.edu
  \IEEEcompsocthanksitem Zhizhong Han is with the Department of Computer Science, Wayne State University, Detroit, USA. E-mail: h312h@wayne.edu
}

\thanks{
The corresponding author is Huifang Feng and Yu-Shen Liu.
The main work of this paper was completed at Tsinghua University.
The source code, data and pretrained models are available at \textcolor{red}{\href{https://github.com/LeoQLi/PFF-Net}{https://github.com/LeoQLi/PFF-Net}}.
}
}

\markboth{Journal of \LaTeX\ Class Files,~Vol.~18, No.~9, September~2020}
{How to Use the IEEEtran \LaTeX \ Templates}

\maketitle

\begin{abstract}
  Estimating the normal of a point requires constructing a local patch to provide center-surrounding context, but determining the appropriate neighborhood size is difficult when dealing with different data or geometries.
  Existing methods commonly employ various parameter-heavy strategies to extract a full feature description from the input patch.
  However, they still have difficulties in accurately and efficiently predicting normals for various point clouds.
  In this work, we present a new idea of feature extraction for robust normal estimation of point clouds.
  We use the fusion of multi-scale features from different neighborhood sizes to address the issue of selecting reasonable patch sizes for various data or geometries.
  We seek to model a patch feature fitting (PFF) based on multi-scale features to approximate the optimal geometric description for normal estimation and implement the approximation process via multi-scale feature aggregation and cross-scale feature compensation.
  The feature aggregation module progressively aggregates the patch features of different scales to the center of the patch and shrinks the patch size by removing points far from the center.
  It not only enables the network to precisely capture the structure characteristic in a wide range, but also describes highly detailed geometries.
  The feature compensation module ensures the reusability of features from earlier layers of large scales and reveals associated information in different patch sizes.
  Our approximation strategy based on aggregating the features of multiple scales enables the model to achieve scale adaptation of varying local patches and deliver the optimal feature description.
  Extensive experiments demonstrate that our method achieves state-of-the-art performance on both synthetic and real-world datasets with fewer network parameters and running time.
\end{abstract}


\section{Introduction}  \label{sec:intro}

\IEEEPARstart{P}{oint} cloud normal estimation is one of the basic tasks in 3D computer vision.
It has a very wide range of applications and is a prerequisite for many downstream tasks or algorithms, such as surface reconstruction~\cite{kazhdan2006poisson}, graphics rendering~\cite{blinn1978simulation,gouraud1971continuous,phong1975illumination}, point cloud denoising~\cite{avron2010L1,lu2017gpf,lu2020low,liu2023pcdnf,de2023contrastive} and so on.
Although normal estimation has been extensively studied with the development of 3D point cloud processing, it is still challenging under varying noise levels, non-uniform sampling densities, and various complex geometries.

\begin{figure}[t]
	\centering
	\includegraphics[width=\linewidth]{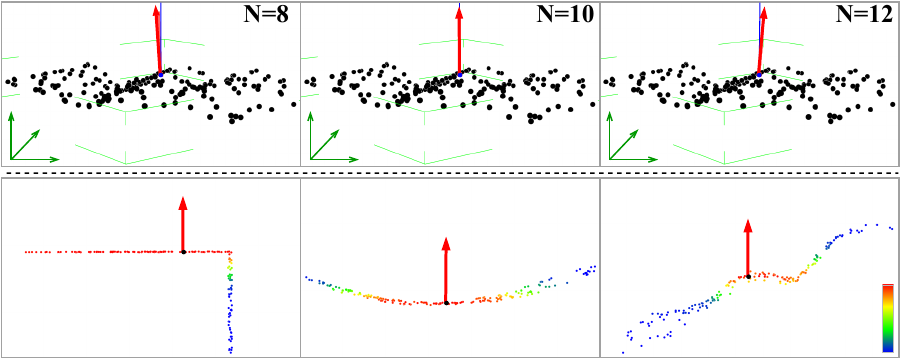} \vspace{-0.7cm}
  \caption{
    Normal estimation for point cloud patches.
    \textit{Top row}: on a 3D surface, the normals estimated by different neighborhood sizes $N$ have different directions.
    \textit{Bottom row}: the number of neighboring points that can be used to accurately estimate the query point normal varies in different structures.
    The red points contribute more for normal estimation, and the blue points contribute less.
  }
  \label{fig:intro}
  \vspace{-0.2cm}
\end{figure}

A standard procedure for estimating a query point normal is to build a fixed-scale local patch and analyze its geometry using various techniques~\cite{hoppe1992surface,guerrero2018pcpnet,ben2020deepfit}.
However, as shown in Fig.~\ref{fig:intro}, it is difficult to choose an appropriate patch size for different data or geometries.
A too small patch size can not provide enough neighboring points to capture the local spatial geometric information, while a too large patch size will bring redundancy or dehighlight sharp geometry, degenerating accuracy and efficiency.
We provide an experimental analysis in Sec.~\ref{sec:pre}.
Specifically,
(1) for point clouds with noise, a relatively larger size is often a better choice.
(2) For structures with smooth planes, it is suitable to select a smaller patch size, and a larger size may not bring performance improvement but redundant points and additional computational burden.
(3) For structures with high curvature, the points that are favorable for normal estimation are basically distributed in a small range around the query point, while the points far from the center are mostly irrelevant.
Therefore, existing learning-based methods~\cite{ben2019nesti,zhu2021adafit,zhou2022refine,li2022hsurf,li2023NeAF,li2023shsnet,xiu2023msecnet} adopt various techniques to extract features from point cloud patches to fully capture the local geometries.

\begin{figure}[t]
  \centering
  \includegraphics[width=\linewidth]{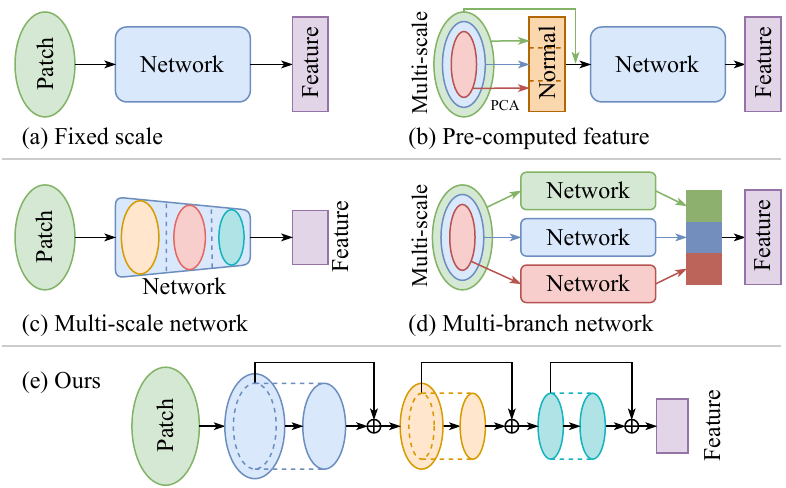} \vspace{-0.7cm}
  \caption{
      Previous methods use (a) fixed-scale patches (DeepFit~\cite{ben2020deepfit}, GraphFit~\cite{li2022graphfit}), (b) pre-computed features (Refine-Net~\cite{zhou2022refine}, Zhang \etal~\cite{zhang2022geometry}), (c) multi-scale networks (AdaFit~\cite{zhu2021adafit}, NeAF~\cite{li2023NeAF}, HSurf-Net~\cite{li2022hsurf}) or (d) multi-branch networks (PCPNet~\cite{guerrero2018pcpnet}, Nesti-Net~\cite{ben2019nesti}, SHS-Net~\cite{li2023shsnet,li2024shsnet-pami}).
      (e) Our patch feature fitting method for normal estimation.
  }
  \label{fig:net_intro}
  \vspace{-0.2cm}
\end{figure}

The key insight of this work is that aggregating features from different neighborhood sizes can address the issue of selecting reasonable patch sizes for various data or geometries.
For noisy point clouds, having the central points acquire information in a large neighborhood can make the estimation more robust.
For smooth planes or large curvature structures, letting the model focus on points near the center can lead to more efficient and accurate normal estimation.
However, how to efficiently learn multi-scale features from input patches and effectively make these features reveal the geometry is not properly solved.
Existing methods shown in Fig.~\ref{fig:net_intro} estimate point normals by first establishing local patches, and then utilizing neural networks to learn to directly or indirectly map their extracted features to 3D normal vectors.
However, the methods in Fig.~\ref{fig:net_intro}(a) do not generalize well to various data.
The methods in Fig.~\ref{fig:net_intro} (b) and (d) suffer from a large number of network parameters or high computational complexity.
The methods in Fig.~\ref{fig:net_intro}(c) have limited feature learning capabilities with reduced points.

\begin{figure*}[t]
	\centering
	\includegraphics[width=\linewidth]{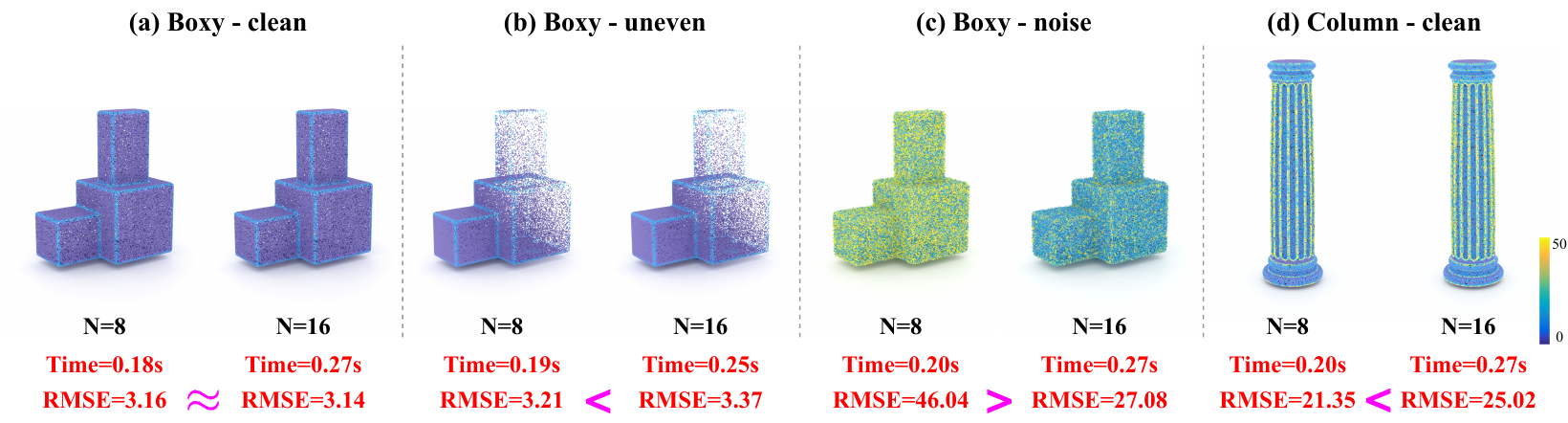} \vspace{-0.7cm}
  \caption{
    Normal estimation results of PCA on different point clouds using two different neighbor sizes $N$.
    The execution time and average normal angle RMSE are provided under each point cloud.
    The point color is the normal angle RMSE mapped to a heatmap ranging from $0^{\circ}$ to $50^{\circ}$.
  }
  \label{fig:PCA_toy}
  \vspace{-0.2cm}
\end{figure*}

As shown in Fig.~\ref{fig:net_intro}(e), our method attempts to model a patch feature fitting based on multi-scale features to approximate the optimal geometric description for point cloud normal estimation.
We implement an approximation process via multi-scale feature aggregation and cross-scale feature compensation.
Unlike surface fitting models~\cite{cazals2005estimating,ben2020deepfit,zhu2021adafit,li2022graphfit,du2023rethinking} that use truncated Taylor expansion to fit 3D surfaces for point cloud patches, we view the feature extraction as a higher-order approximation process in feature space using neural networks.
The approximation process is built using multi-scale features from different patch sizes, and the approximation error is considered via cross-scale feature compensation based on attention.
Instead of treating all input points equally, we collect more features from points closer to the center of the patch, and transfer features from large scales to small scales for multi-scale feature aggregation.
Thus, the model can not only efficiently capture a wide range of spatial information, but also perpetually focus on central points.
Our approximation strategy based on weighing the features from multiple scales enables the model to achieve scale adaptation for various geometries.
Experimental results on the shape dataset, the real-world indoor and outdoor scene datasets show that our method is robust to domain shift (training on shapes, testing on scenes) and has good generalization capability on real-world LiDAR data.
We also demonstrate the extensibility of our method and its ability to improve the performance of other methods.
Our main contributions can be summarized as follows.
\begin{itemize}[leftmargin=*]
\setlength{\itemsep}{0pt}
\setlength{\parsep}{0pt}
\setlength{\parskip}{0pt}
    \item We propose a strategy that exploits the idea of patch feature fitting based on multi-scale features to approximate the optimal features for normal estimation.
    \item We design an effective network architecture that includes multi-scale feature aggregation and cross-scale feature compensation to implement the patch feature fitting and the error in approximation process.
    \item Extensive evaluation shows that our strategy brings significant performance improvements with fewer parameters and runtime than baseline methods.
\end{itemize}

\section{Related Work}

\noindent
\textbf{Traditional approaches}.
The most widely used normal estimation method is based on the classic Principle Component Analysis (PCA)~\cite{hoppe1992surface}, which analyzes the variance in a patch around a query point and defines its normal as the direction of minimal variance~\cite{stewart1993early}.
Later, many improvements~\cite{alexa2001point,lange2005anisotropic,huang2009consolidation} have been proposed for PCA.
Mitra \etal~\cite{mitra2003estimating} costly investigate the effect of local curvature and point density of the underlying surface to determine the patch size.
To preserve more detailed features, other methods use Voronoi cells~\cite{amenta1999surface,merigot2010voronoi,dey2006provable,alliez2007voronoi}, Hough transform~\cite{boulch2012fast} and edge-aware sampling~\cite{huang2013edge}.
Variants that are based on complex surfaces have also been proposed, such as moving least squares~\cite{levin1998approximation}, jet fitting~\cite{cazals2005estimating}, spherical fitting~\cite{guennebaud2007algebraic}, multi-scale kernel~\cite{aroudj2017visibility}, local kernel regression~\cite{oztireli2009feature} and winding-number field~\cite{xu2023globally}.
The data-specific parameters of the above methods often do not generalize well to different data.

\noindent
\textbf{Learning-based approaches}.
(1) \emph{Regression-based methods}.
In recent years, learning-based methods have been proposed to directly predict normals from point clouds in a data-driven manner.
Some methods~\cite{boulch2016deep,roveri2018pointpronets,lu2020deep} try to map the unstructured point cloud data into a regular domain to extract features.
Meanwhile, some alternative approaches learn from raw point clouds.
PCPNet~\cite{guerrero2018pcpnet} and Zhou \etal~\cite{zhou2020normal} adopt the PointNet architecture~\cite{qi2017pointnet} to extract patch features from multiple scales.
Hashimoto \etal~\cite{hashimoto2019normal} use a two-branch network to extract local and spatial features.
Nesti-Net~\cite{ben2019nesti} tries to search the optimal neighborhood scale but suffers from computational inefficiency.
Refine-Net~\cite{zhou2022refine} employs a refinement network to optimize the initial normals using the learned local features.
HSurf-Net~\cite{li2022hsurf} introduces a hyper surface that is parameterized by MLP layers and optimized in high dimensional feature space for normal estimation.
NeAF~\cite{li2023NeAF} proposes to implicitly learn the angle distance field of points and predict the angle offsets of query vectors.
SHS-Net~\cite{li2023shsnet,li2024shsnet-pami} introduces signed hyper surface to learn unoriented and oriented normals.
NGLO~\cite{li2023neural} first predicts coarse normals with global consistency from the global point cloud by learning implicit functions, and then refines the normals based on local information to improve their accuracy.
NeuralGF~\cite{li2023neuralgf} estimates normals in an unsupervised manner by learning gradients of implicit functions.
MSECNet~\cite{xiu2023msecnet} improves normal estimation in areas with drastic normal changes by introducing edge detection technology.
CMG-Net~\cite{wu2024cmg} proposes a metric of Chamfer Normal Distance to address the issue of normal direction inconsistency in noisy point clouds.
(2) \emph{Fitting-based methods}.
Lenssen \etal~\cite{lenssen2020deep} propose to iteratively parameterize an adaptive anisotropic kernel to learn weights for a least squares plane fitting.
MTRNet~\cite{cao2021latent} uses a differentiable RANSAC to fit a latent tangent plane.
DeepFit~\cite{ben2020deepfit}, AdaFit~\cite{zhu2021adafit}, GraphFit~\cite{li2022graphfit}, Zhou \etal~\cite{zhou2023improvement}, Zhang \etal~\cite{zhang2022geometry} and Du \etal~\cite{du2023rethinking} use a PointNet-like or graph convolutional network to predict point weights and apply a weighted polynomial surface fitting to fit a local surface and calculate its normal.
They try to use the weight to provide more reliable inlier points for the fitting.
The above methods do not fully explore multi-scale features and effectively filter out redundant information, often have parameter-heavy networks and run inefficiently.
Compared to AdaFit, which reduces patch size to simplify explicit surface fitting, our center-based downsampling is designed to enrich geometric representation across scales for direct normal regression. By integrating point-wise weighting, multi-mode feature fusion, and cross-scale compensation, our method achieves higher accuracy with fewer parameters and significantly better efficiency.

\section{Preliminary}  \label{sec:pre}

\subsection{Effect of patch size on normal estimation}

To analyze the effect of different patch sizes on point cloud normal estimation, we employ the classic PCA algorithm~\cite{hoppe1992surface} with k-nearest neighbor sizes $N\!=\!8$ and $N\!=\!16$ to conduct normal calculations on several different point clouds, which cover the common situations including clean (noise-free), noise, non-uniform sampling, simple and complex structures.
The results are shown in Fig.~\ref{fig:PCA_toy}, we can observe several phenomenons that are in harmony with the intuition:
(1) Smaller patch size takes less time, while larger one takes more time.
(2) For simple flat structures (boxy shape), the results of different patch sizes are similar, but a larger size leads to redundant information and costs much more runtime.
(3) For complex structures (column shape), the useful information for the query point normal estimation is relatively concentrated.
A smaller size brings better results, while a larger size gives invalid information or even distractions.
(4) For non-uniformly sampled point clouds, a smaller patch size can provide more reasonable structural information from points that are far apart.
(5) For noisy point clouds, a larger patch size can effectively suppress the interference caused by noise.
This experiment shows that using a large or small patch size has advantages and disadvantages in different situations, and the optimal method should take the advantages of both and avoid the disadvantages of both.
To this end, we propose a strategy that exploits the idea of patch feature fitting based on multi-scale features to approximate the optimal features and enables the model to achieve scale adaptation for various geometries in normal estimation.
We use a relatively large patch size to extract reliable structural features and suppress noise, and gradually reduce the patch size to focus on the center of the patch and get faster runtime.
Meanwhile, the feature aggregation during scale reduction filters out redundant or invalid information to ensure accurate normal estimation.

\subsection{Theoretical motivation from Taylor expansion}

Given a point cloud patch $P \!=\! \{ p_i\}_{i=1}^N$ consisting of $N$ points around a query point $q$, our method aims to estimate the unoriented normal $\mathbf{n}_{q}$ at point $q$.
Our network design is inspired by the Taylor expansion of an implicit surface function.
Let $f : \mathbb{R}^3 \rightarrow \mathbb{R}$ be a smooth scalar function that defines a surface by $f(p) = 0$, where $\nabla f(p)$ is aligned with the surface normal at $p$.
For any point $p_i$ in the neighborhood of $q$, the function can be locally approximated via a second-order Taylor expansion:
\begin{equation}  \label{eq:taylor}
  f(q + X) = f(q) + \nabla f(q)^\top X + \frac{1}{2} X^\top H_f(q) X + R_3(X) ~,
\end{equation}
where $X \!=\! p_i - q$ denotes the local offset, $H_f(q)$ is the Hessian matrix capturing curvature, and $R_3(X)$ represents the third- and higher-order residual.
The first two terms $\nabla f(q)^\top X$ and $X^\top H_f(q) X$ correspond to local planar and quadratic geometry, while the residual term accounts for finer surface details.
In practice, the surface fitting based methods~\cite{ben2020deepfit,zhu2021adafit,li2022graphfit} first approximate a 3D surface by a binary polynomial and then compute the normal of the fitted local surface using the coefficient of the solved polynomial.

In order to directly regress 3D point normals in an end-to-end manner, we approximate the surface function in a feature space and define a learnable feature transformation $F(X)$ based on the offset $X$.
We expect $F(X)$ to capture geometric information such as normals and curvatures.
As a result, we view $F(X)$ as a feature-space analogue of a polynomial expansion:
\begin{equation}  \label{eq:poly}
  F(X) \approx \theta_1^\top X + X^\top \theta_2 X + \cdots ~,
\end{equation}
where $\theta_1$ and $\theta_2$ are learnable parameters of MLPs approximating surface derivatives, and the terms correspond to surface differential properties.
As introduced in Sec.~\ref{sec:conv}, we interpret our network as a functional approximation of this expansion, with each residual block modeling a distinct component of the underlying polynomial that characterizes the local surface geometry.
Specifically, block $\mathcal{F}_1$ is designed to progressively extract and aggregate multi-scale features $\mathcal{X}$ corresponding to the linear and quadratic terms.
Each residual block in $\mathcal{F}_1$ updates the representation as
\begin{equation}
  \mathcal{X}_{k+1} = \mathcal{X}_k + \text{MLP}_k \left( \mathcal{X}_k, X_k \right) ~,
\end{equation}
which can be viewed as fitting the dominant geometric components $\theta_1^\top X$ and $X^\top \theta_2 X$ through stage-wise feature refinement.
$X_k$ is also used to denote offset encoding.
Each $\text{MLP}_k(\cdot)$ is responsible for modeling one stage of the expansion, and the residual addition naturally mimics term-wise accumulation.
In effect, $\mathcal{F}_1$ fits a local planar approximation plus curvature, \ie, the first two low-order terms of the Taylor expansion.
This is analogous to classical geometry-fitting: estimating the tangent plane (first-order) and principal curvatures (second-order) via a truncated Taylor expansion (called $n$-jet)~\cite{cazals2005estimating}.

Subsequently, block $\mathcal{F}_2$ models the remaining high-order variation using finer-scale patches:
\begin{equation}
  \mathcal{X}_{\text{refine}} = \mathcal{X}_{\text{coarse}} + \text{MLP}_{\text{refine}}(\mathcal{X}_{\text{coarse}}) ~,
\end{equation}
serving as a learned residual corrector to capture detailed geometry in complex regions.
Additionally, our attention-based cross-scale compensation module in Sec.~\ref{sec:attn} reuses coarse-scale features in a weighted manner, further enhancing the fidelity of the approximation.

In summary, our network mimics the progressive accumulation of Taylor terms: block $\mathcal{F}_1$ fits the low-order structure, block $\mathcal{F}_2$ corrects the residual, and cross-scale attention enables explicit compensation.
The formulation not only grounds our architecture in geometric theory but also explains the effectiveness of our residual-based multi-scale feature fusion.

\section{Method}

\begin{figure*}[t]
	\centering
	\includegraphics[width=.95\linewidth]{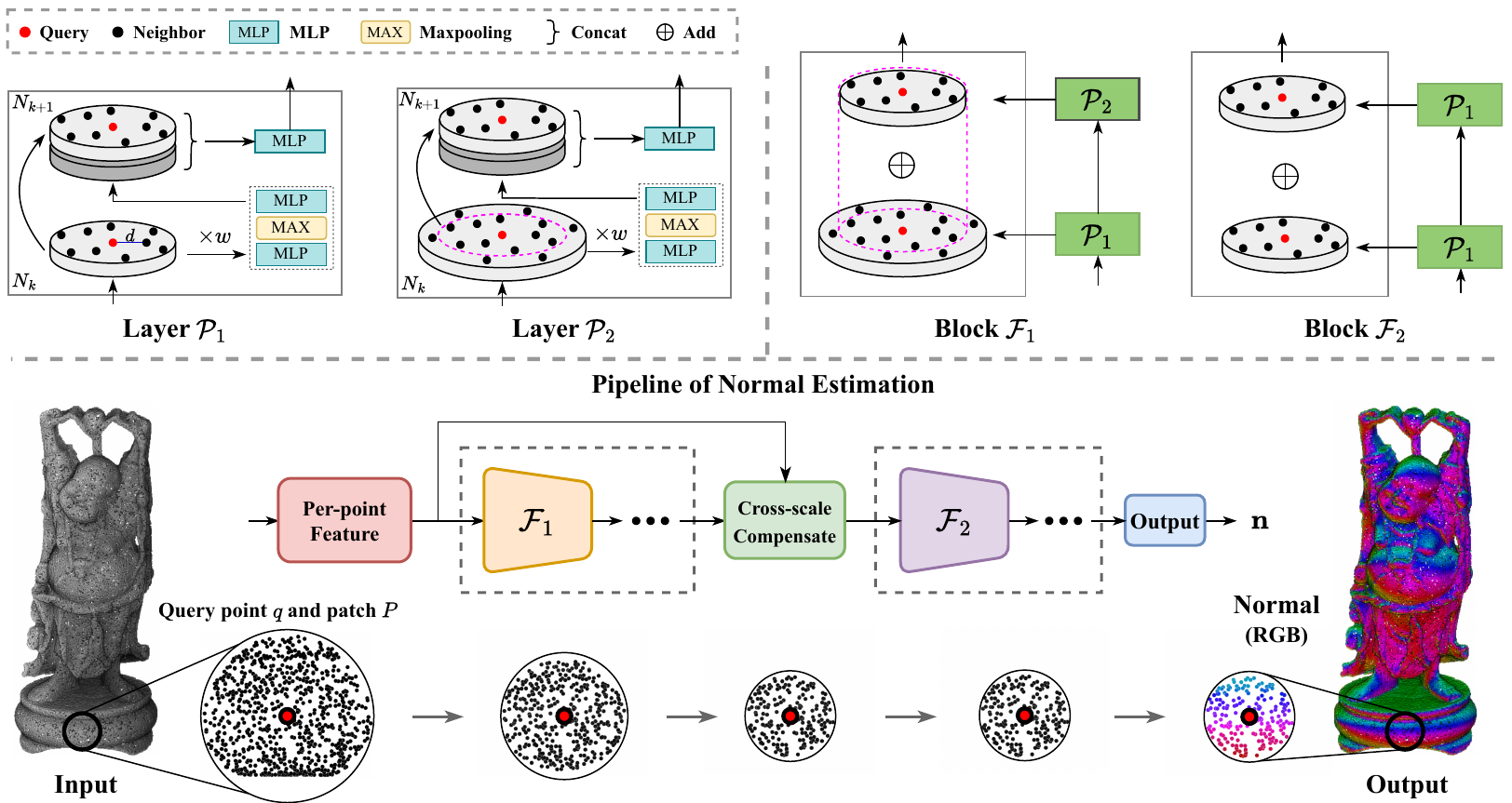}  \vspace{-0.15cm}
  \caption{
    An overview of our normal estimation method.
    The blocks $\mathcal{F}_1$ and $\mathcal{F}_2$ are built by using different layers $\mathcal{P}_1$ and $\mathcal{P}_2$.
    Two blocks are stacked recursively to form the normal estimation pipeline.
    The color of the shape point cloud indicates the predicted normals.
  }
  \label{fig:net}
\end{figure*}

\noindent\textbf{Overview}.
In this work, we introduce a novel multi-scale feature extraction method in normal estimation.
Rather than explicitly fitting 3D surfaces using point coordinates, we implement a feature extraction and fusion mechanism based on multiple size scales to allow the network to adaptively find optimal geometric descriptions for input patches with fixed scales from their fused features.
Our goal is to use the aggregation of multi-scale features to obtain $F(X)$, which is implicitly defined by the ground truth normal.
To this end, we design two kinds of layers with different learning strategies, which are further used to build two kinds of blocks, to extract features at different scales.
Moreover, we employ an attention block to provide compensation concerning the approximation error.
Fig.~\ref{fig:net} shows an overview of the proposed algorithm, which mainly consists of a per-point feature extraction module, two blocks for multi-scale feature aggregation and a cross-scale compensation module using attention mechanism.
As the number of layers in the network increases, we decrease the patch size in the latter layers by reducing the nearest neighbors of the query $q$.
The reduction in points effectively reduces the burden on the algorithm and brings higher running efficiency.

\begin{figure}[t]
	\centering
	\includegraphics[width=\linewidth]{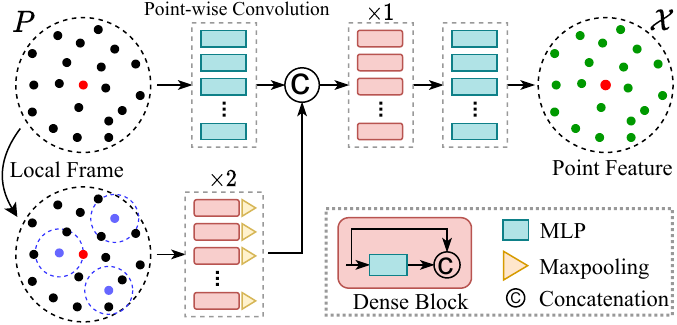}  \vspace{-0.5cm}
  \caption{
    The architecture of our per-point feature extraction module.
  }
  \label{fig:feat_extra}
  \vspace{-0.2cm}
\end{figure}

\subsection{Per-point Feature Extraction}  \label{sec:feat}

We first learn a point-wise feature set $\mathcal{X} \!=\! \{ x_i\}_{i=1}^N$ for all points of the input patch $P$.
Previous methods~\cite{guerrero2018pcpnet,ben2020deepfit,zhu2021adafit,li2023NeAF} employ the PointNet-like structures~\cite{qi2017pointnet} for feature extraction.
However, their network is insufficient to capture local structure information since it does not encode the connections of each point to its neighborhoods.
As shown in Fig.~\ref{fig:feat_extra}, we provide a novel feature extraction unit, which is formulated as
\begin{equation}  \label{eq:unit}
  x_i = \psi \left( \phi(p_i), ~\text{MAX} \big\{\varphi(p_i^j|j=1,\!\cdots\!,n_k) \big\} \right), ~i\!=\!1,\!\cdots\!,N,
\end{equation}
where $\psi$ and $\phi$ are MLPs, and $\varphi$ is a stack of densely connected graph convolution layers. $p_i^j \!\in\! \text{kNN}(p_i)$ denotes the $n_k$-nearest neighboring points of $p_i$.
$\text{MAX}\{\cdot\}$ means maxpooling.
The function $\psi(\cdot,\cdot)$ fuses the point-wise features and local features for each point by concatenation.
The graph convolution extracts local features for each point in the patch, while the dense connection delivers features with richer contextual information~\cite{huang2017densely, liu2019densepoint}.

\subsection{Multi-scale Feature Aggregation}  \label{sec:conv}

\noindent\textbf{A template}.
To construct the feature-based polynomial, we learn the multi-scale feature based on the per-point feature $\mathcal{X}$.
Specifically, we extract features from different patch sizes and then fuse the features to realize feature aggregation.
During this process, we gradually reduce the patch size around the query point $q$ by removing neighboring points that are far away from it, and aggregate the information from the removed far points to the remaining points in the patch, thus realizing focalization and concentration.
Meanwhile, the reduction in points can improve running efficiency.
In summary, the basic convolutional layer $\mathcal{P}$ is formulated as
\begin{equation}  \label{eq:layer}
    y_{i} \!=\! \alpha \left(\beta \left(\text{MAX} \big\{ \gamma (w_{j} \cdot x_{j}) \big\}_{j=1}^{N_{k}} \right), ~x_{i} \right), i\!=\!1,\!\cdots\!,N_{k+1} ~,
\end{equation}
where $\alpha, \beta$ and $\gamma$ are MLPs, $x_{i}$ and $y_{i}$ are per-point features.
Note that the features are sorted in ascending order according to the distances between the underlying points and the query, so that the index can be used to efficiently remove points and their features that are far away.
$k$ is the size index and $N_{k+1} \leqslant N_{k} \leqslant N$.
$\text{MAX}\{\cdot\}$ represents the feature maxpooling over $N_{k}$ points in the patch.
$w$ is a distance-based weight~\cite{li2023shsnet,li2024shsnet-pami}, and is calculated by
\begin{equation}  \label{eq:weight}
    w_{j} = \frac{d_j}{\sum_{i=1}^{N}{d_i}}, ~ d_i = \sigma \big(a - b{||p_i - q||}_2 \big),
\end{equation}
where $\sigma$ is the sigmoid function, $a$ and $b$ are learnable parameters with the initial value set to $1$.
We use the weight $w$ to let the network focus more on the points $p_i$ that are closer to the query point $q$, thereby extracting reliable features in areas where the geometry changes drastically and improving the robustness of the algorithm.
Although our feature aggregation framework shares conceptual similarities with SHS-Net~\cite{li2023shsnet}, it is fundamentally different in motivation and scope of feature encoding.
SHS-Net adopts a simplified and lightweight fusion of global and local features, while our approach is rooted in Taylor expansion theory and performs hierarchical residual fusion solely within local neighborhoods, enabling finer geometric approximation.

\noindent\textbf{Two variants}.
Next, we will build two different layers and two different blocks based on Eq.~\eqref{eq:layer}.
As shown in Fig.~\ref{fig:net}, we provide two types of layers $\mathcal{P}_1$ and $\mathcal{P}_2$ depending on whether the patch size is reduced or not, \ie, $N_{k+1} \!=\! N_{k}$ or $N_{k+1} \!<\! N_{k}$.
These two layers $\mathcal{P}_1$ and $\mathcal{P}_2$ can be stacked alternately, enabling the model to find increasingly rich representations of the point cloud patch.
In our normal estimation pipeline, we further build two different blocks $\mathcal{F}_1$ and $\mathcal{F}_2$ using the different combinations of layers $\mathcal{P}_1$ and $\mathcal{P}_2$.
For block $\mathcal{F}_1$, the features from two layers $\mathcal{P}_1$ and $\mathcal{P}_2$ are aggregated by addition operation and passed to the next layer, \ie,
\begin{equation}
  \mathcal{X}_{k+1} = [\mathcal{X}_k]_{N_{k+1}} + \mathcal{P}_2(\mathcal{X}_k), ~~ \mathcal{X}_{k} = \mathcal{P}_1(\mathcal{X}_{k-1}),
\end{equation}
where $[\cdot]_{N_{k+1}}$ means taking the features of the nearest neighbor point of $q$ whose neighborhood size is ${N_{k+1}}$.
Note that we sort the input points and their features according to their 3D distance from $q$ and keep the order unchanged during processing, thus achieving fast indexing.
For block $\mathcal{F}_2$, the features from two $\mathcal{P}_1$ layers are aggregated by addition operation and then passed to the next layer, \ie,
\begin{equation}
  \mathcal{X}_{k+1} = \mathcal{X}_k + \mathcal{P}_1(\mathcal{X}_k), ~~ \mathcal{X}_{k} = \mathcal{P}_1(\mathcal{X}_{k-1}),
\end{equation}
The block $\mathcal{F}_1$ extracts wide-range neighborhood features and gradually reduces the patch size to filter out redundant or irrelevant features, while the block $\mathcal{F}_2$ further refines these features.
With the recursive utilization of our blocks, the large patch sizes of earlier layers provide more information about the underlying geometries, while the small patch sizes of the latter layers result in a more accurate description of the central details.
By utilizing their combination, we can use fewer network parameters to extract information useful for normal estimation, and the reduction in patch size can also improve the algorithm's running efficiency.

\begin{table*}[t]
\centering
\footnotesize
\setlength{\tabcolsep}{1.6mm}
\caption{
	Normal RMSE on the datasets PCPNet and FamousShape.
	The lower the better.
}
\vspace{-0.25cm}
\label{table:pcpnet_famousShape}
\begin{tabular}{l|cccc|cc| >{\columncolor{mygray}} c||  cccc|cc| >{\columncolor{mygray}} c}
	\toprule
	\multirow{3}{*}{Category} & \multicolumn{7}{c||}{\textbf{PCPNet Dataset}}   & \multicolumn{7}{c}{\textbf{FamousShape Dataset}} \\
	\cmidrule(r){2-15}
	& \multicolumn{4}{c|}{Noise} & \multicolumn{2}{c|}{Density} &
	& \multicolumn{4}{c|}{Noise} & \multicolumn{2}{c|}{Density} &    \\
	& None & Low & Medium & High  & Stripe & Gradient  & \multirow{-2}{*}{{Average}}
	& None & Low & Medium & High  & Stripe & Gradient  & \multirow{-2}{*}{{Average}}    \\
	\midrule
	PCV~\cite{zhang2018multi} 	                 & 12.50 & 13.99 & 18.90 & 28.51 & 13.08 & 13.59 &  16.76    & 21.82 & 22.20 & 31.61 & 46.13 & 20.49 & 19.88 &   27.02  \\
	Jet~\cite{cazals2005estimating}              & 12.35 & 12.84 & 18.33 & 27.68 & 13.39 & 13.13 &  16.29    & 20.11 & 20.57 & 31.34 & 45.19 & 18.82 & 18.69 &   25.79  \\
	PCA~\cite{hoppe1992surface} 	             & 12.29 & 12.87 & 18.38 & 27.52 & 13.66 & 12.81 &  16.25    & 19.90 & 20.60 & 31.33 & 45.00 & 19.84 & 18.54 &   25.87  \\
	PCPNet~\cite{guerrero2018pcpnet}             & 9.64  & 11.51 & 18.27 & 22.84 & 11.73 & 13.46 &  14.58    & 18.47 & 21.07 & 32.60 & 39.93 & 18.14 & 19.50 &   24.95  \\
	Zhou \etal~\cite{zhou2020normal}             & 8.67  & 10.49 & 17.62 & 24.14 & 10.29 & 10.66 &  13.62    & -    & -     & -    & -    & -    & -    & -    \\
	NeuralGF~\cite{li2023neuralgf}               & 7.89  & 9.85  & 18.62 & 24.89 & 9.21  & 9.29  &  13.29    & 13.74 & 16.51 & 31.05 & 40.68 & 13.95 & 13.17 &   21.52  \\
	Nesti-Net~\cite{ben2019nesti}                & 7.06  & 10.24 & 17.77 & 22.31 & 8.64  & 8.95  &  12.49    & 11.60 & 16.80 & 31.61 & 39.22 & 12.33 & 11.77 &   20.55  \\
	Lenssen \etal~\cite{lenssen2020deep}         & 6.72  & 9.95  & 17.18 & 21.96 & 7.73  & 7.51  &  11.84    & 11.62 & 16.97 & 30.62 & 39.43 & 11.21 & 10.76 &   20.10  \\
	DeepFit~\cite{ben2020deepfit}                & 6.51  & 9.21  & 16.73 & 23.12 & 7.92  & 7.31  &  11.80    & 11.21 & 16.39 & 29.84 & 39.95 & 11.84 & 10.54 &   19.96  \\
	MTRNet~\cite{cao2021latent}                  & 6.43  & 9.69  & 17.08 & 22.23 & 8.39  & 6.89  &  11.78    & -    & -     & -     & -     & -    & -    & -    \\
	Refine-Net~\cite{zhou2022refine}             & 5.92  & 9.04  & 16.52 & 22.19 & 7.70  & 7.20  &  11.43    & -    & -     & -     & -     & -    & -    & -    \\
	Zhang \etal~\cite{zhang2022geometry}         & 5.65  & 9.19  & 16.78 & 22.93 & 6.68  & 6.29  &  11.25    & 9.83 & 16.13 & 29.81 & 39.81 & 9.72 & 9.19 &   19.08  \\
	Zhou \etal~\cite{zhou2023improvement}        & 5.90  & 9.10  & 16.50 & 22.08 & 6.79  & 6.40  &  11.13    & -    & -     & -     & -     & -    & -    & -    \\
	AdaFit~\cite{zhu2021adafit}                  & 5.19  & 9.05  & 16.45 & 21.94 & 6.01  & 5.90  &  10.76    & 9.09 & 15.78 & 29.78 & 38.74 & 8.52 & 8.57 &   18.41  \\
	GraphFit~\cite{li2022graphfit}               & 5.21  & 8.96  & 16.12 & 21.71 & 6.30  & 5.86  &  10.69    & 8.91 & 15.73 & 29.37 & 38.67 & 9.10 & 8.62 &   18.40  \\
	NeAF~\cite{li2023NeAF}                       & 4.20  & 9.25  & 16.35 & 21.74 & 4.89  & 4.88  &  10.22    & 7.67 & 15.67 & 29.75 & 38.76 & 7.22 & 7.47 &   17.76  \\
	HSurf-Net~\cite{li2022hsurf}                 & 4.17  & 8.78  & 16.25 & 21.61 & 4.98  & 4.86  &  10.11    & 7.59 & 15.64 & 29.43 & 38.54 & 7.63 & 7.40 &   17.70  \\
	NGLO~\cite{li2023neural}  	                 & 4.06  & 8.70  & 16.12 & 21.65 & 4.80  & 4.56  &  9.98     & 7.25 & 15.60 & 29.35 & 38.74 & 7.60 & 7.20 &   17.62   \\
    Du \etal~\cite{du2023rethinking}             & 3.85  & 8.67  & 16.11 & 21.75 & 4.78  & 4.63  &  9.96     & 6.92 & 15.05 & 29.49 & 38.73 & 7.19 & 6.92 &   17.38  \\
	SHS-Net~\cite{li2023shsnet}  	             & 3.95  & 8.55  & 16.13 & 21.53 & 4.91  & 4.67  &  9.96     & 7.41 & 15.34 & 29.33 & 38.56 & 7.74 & 7.28 &   17.61   \\
	CMG-Net~\cite{wu2024cmg}                     & 3.87  & 8.45  & 16.08 & 21.89 & 4.85  & 4.45  &  9.93     & 7.07 & 14.83 & 29.04 & 38.93 & 7.43 & 7.03 &   17.39  \\
	MSECNet~\cite{xiu2023msecnet}                & 3.84  & 8.74  & 16.10 & 21.05 & 4.34  & 4.51  &  9.76     & 6.85 & 15.60 & 29.22 & \textbf{38.13} & \textbf{6.64} & 6.65 &   17.18  \\
	Ours                                         & \textbf{3.32} & \textbf{8.34}  & \textbf{15.63} & \textbf{20.94} & \textbf{4.10} & \textbf{3.92} &   \textbf{9.38}
		                                         & \textbf{6.60} & \textbf{14.68} & \textbf{28.86} &        {38.27} &        {6.86} & \textbf{6.41} &   \textbf{16.95}  \\
	\bottomrule
\end{tabular} 
\vspace{-0.2cm}
\end{table*}

\subsection{Cross-scale Compensation}  \label{sec:attn}

Generally, there is information loss during the feature extraction process, which leads to description errors and affects feature approximation.
The residual connection~\cite{he2016deep} that reuses the features from earlier layers can stabilize training and convergence.
In this work, we introduce an alternative method and propose to weigh features at different scales via distance-weighted attention, and generate compensation in geometric description.
First, we generate several feature components through the following formulas
\begin{equation}
  Q_i = {\theta}_{Q} (y_i),  K_i = {\theta}_{K} (x_i),  V_i = {\theta}_{V} (x_i),  {\Delta}_i = {\theta}_{\Delta} ({Q}_i + {K}_i),
\end{equation}
where $\theta$ is MLP, $x_i \!\in\! \mathcal{X}$ is the per-point feature before block $\mathcal{F}_1$ and $y_i \!\in\! \mathcal{Y}$ is the feature after block $\mathcal{F}_1$.
Then, the cross-scale feature compensation is implemented by an attention block conditioned on $\mathcal{Y}$, that is
\begin{equation}  \label{eq:atten}
  z_i = \eta \left(w_i \cdot V_i \cdot \mu ({\Delta}_i), ~~ y_i \right), ~ i\!=\!1,\!\cdots\!,N_k ~,
\end{equation}
where $\eta$ and $\mu$ are MLPs.
$\mu$ provides attention weights to modulate individual features, while $\eta(\cdot,\cdot)$ fuses two kinds of features through concatenation.
$w$ is the distance-based weight in Eq.~\eqref{eq:weight}.
We do not use a \emph{softmax} operation for normalization as in \cite{zhao2021point,van2022revealing} since it does not bring performance gains, which is verified by ablation experiments.

\begin{figure}[t]
  \centering
  \includegraphics[width=\linewidth]{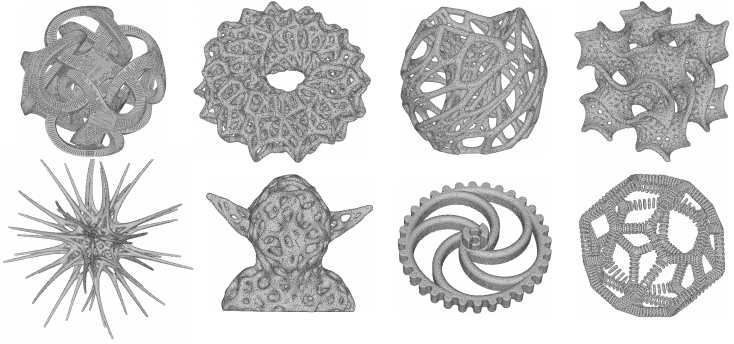}  \vspace{-0.8cm}
  \caption{
    Examples of point cloud shapes from our NestPC dataset.
  }
  \label{fig:NestPC}
  \vspace{-0.2cm}
\end{figure}

\begin{figure*}[t]
  \centering
  \includegraphics[width=\linewidth]{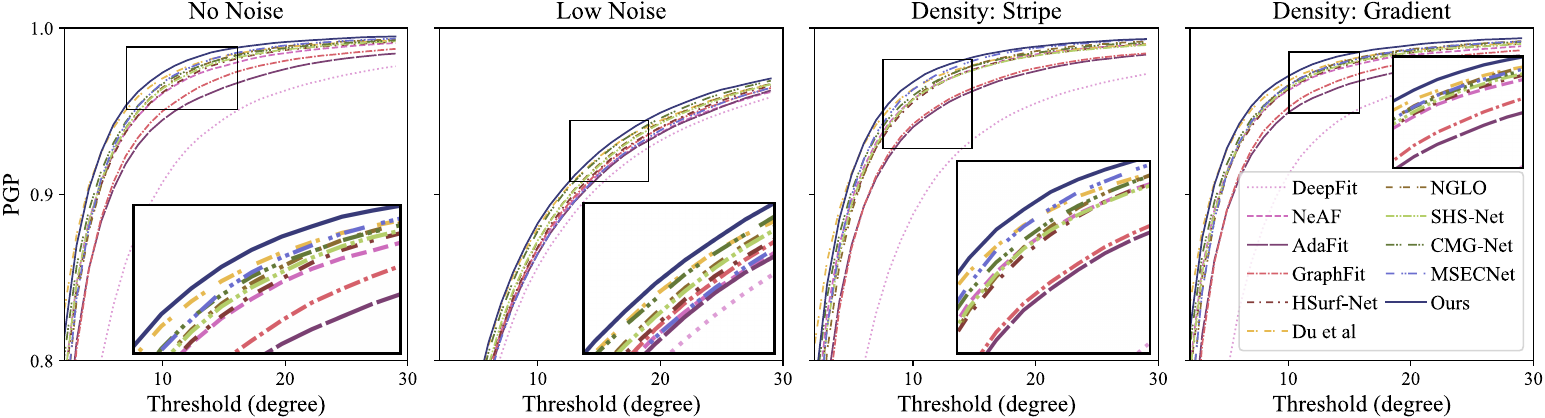}  \vspace{-0.5cm}
  \caption{
    Normal PGP curves on the PCPNet dataset.
    The Y-axis is the percentage of good point normals whose normal errors are smaller than the given angle threshold of the X-axis (in degrees).
    More graphs with a different axis scale are shown in Fig.~\ref{fig:PGP_pcpnet_2}.
  }
  \label{fig:PGP_pcpnet_4}
  \vspace{-0.2cm}
\end{figure*}

\begin{figure}[t]
  \centering
  \includegraphics[width=\linewidth]{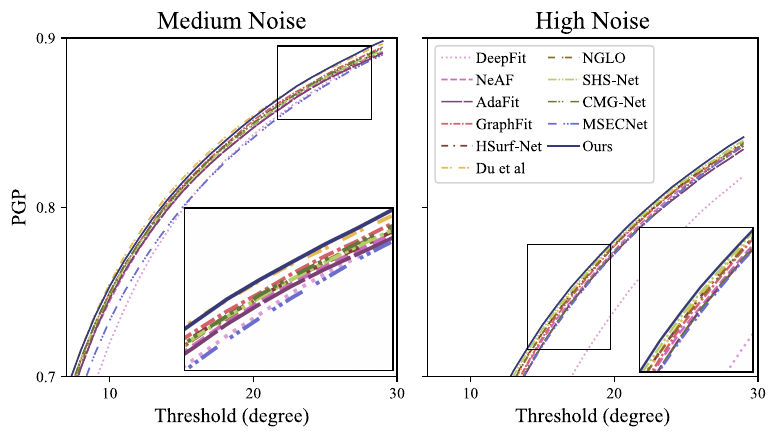}  \vspace{-0.7cm}
  \caption{
    Normal PGP curves on the PCPNet dataset.
    The scale of the axes is different from Fig.~\ref{fig:PGP_pcpnet_4}, so we show it separately for better presentation.
  }
  \label{fig:PGP_pcpnet_2}
  \vspace{-0.2cm}
\end{figure}

\begin{figure}[t]
  \centering
  \includegraphics[width=\linewidth]{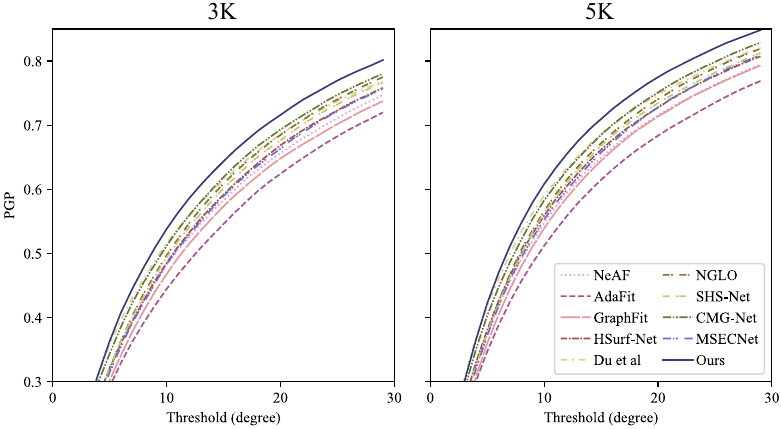}  \vspace{-0.7cm}
  \caption{
    Normal PGP curves on sparse point cloud data with 3000 and 5000 points, respectively.
  }
  \label{fig:PGP_sparse}
  \vspace{-0.2cm}
\end{figure}

\begin{figure}[t]
  \centering
  \includegraphics[width=\linewidth]{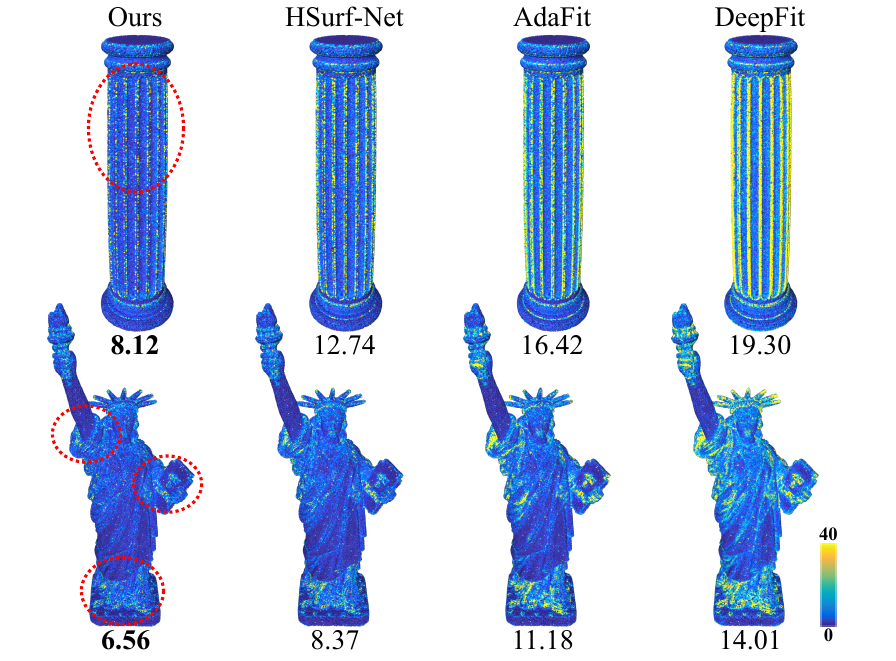}  \vspace{-0.7cm}
  \caption{
    Normal error visualization on complex shapes of the PCPNet dataset.
    The normal RMSE is mapped to a heatmap ($0^{\circ}-40^{\circ}$), and the average value is provided under each shape.
  }
  \label{fig:error_pcpnet}
  \vspace{-0.2cm}
\end{figure}

\begin{figure}[t]
  \centering
  \includegraphics[width=\linewidth]{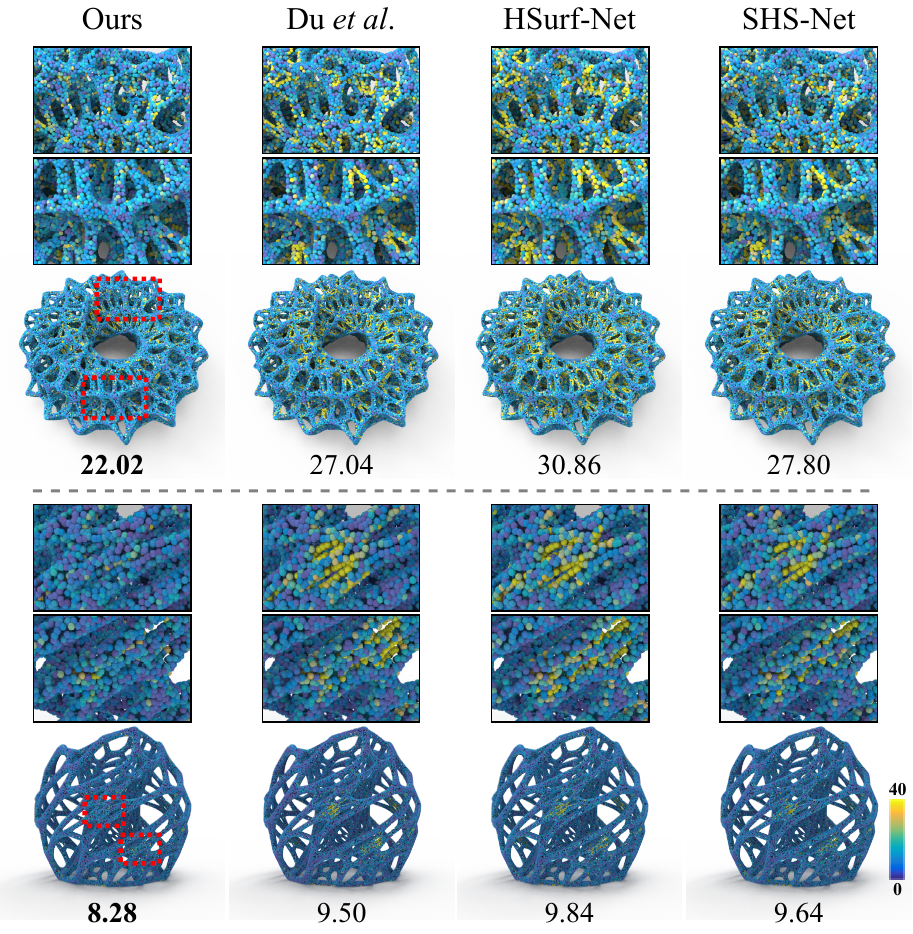}  \vspace{-0.7cm}
  \caption{
    Visual comparison of normal errors on two shapes of the NestPC dataset.
    The normal RMSE is mapped to a heatmap, and the average value is provided under each shape.
  }
  \label{fig:errorMap}
  \vspace{-0.2cm}
\end{figure}

\begin{figure}[t]
  \centering
  \footnotesize
  \includegraphics[width=\linewidth]{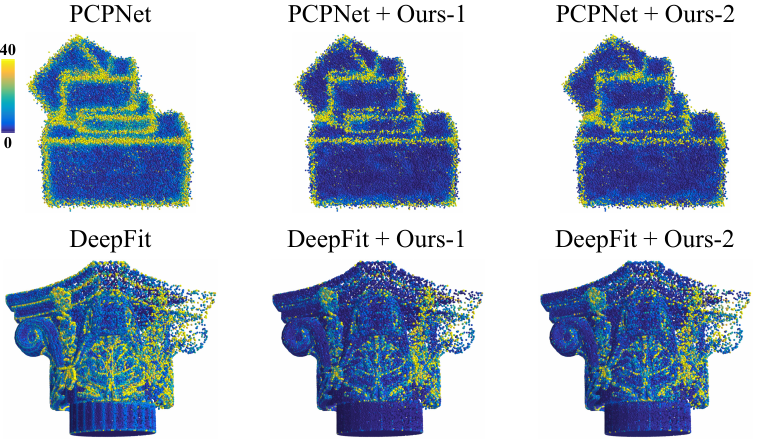}  \vspace{-0.6cm}
  \caption{
    Visual comparison of normal errors on noisy and non-uniform point clouds.
    Our algorithm can be used to improve other methods.
  }
  \label{fig:extension}
  \vspace{-0.2cm}
\end{figure}

\subsection{Normal Prediction and Training Loss}

After obtaining the final output feature $\mathcal{X}_o$ of the remaining nearest $N_o$ points using block $\mathcal{F}_2$, the unnormalized normal of query point $\mathbf{n}_q$ is predicted by a weighted maxpooling of its neighboring features $x_{i} \!\in\! \mathcal{X}_o$, \ie,
\begin{equation}  \label{eq:output}
  \mathbf{n}_q = \delta \big(\text{MAX} \{ w_i \cdot \tau_i \cdot x_{i} | i\!=\!1,\cdots,N_o \} \big) ,
\end{equation}
where $\tau_i \!=\! \text{sigmoid}(\xi(x_{i}))$ is a weight. $\delta$ and $\xi$ are MLPs.
The neighboring point normals $\mathbf{n}_{i}$ are predicted from $\mathcal{X}_o$ by another MLP.
To train the network to predict accurate normal, we calculate the $sin$ distance $d_{\rm sin}$ and squared Euclidean distance $d_{\rm euc}$ between the predicted normal $\mathbf{n}$ and the ground truth $\hat{\mathbf{n}}$, then we have
\begin{equation}  \label{eq:loss_n}
  \mathcal{L}_{\mathbf{n}} = d_{\rm sin} + d_{\rm euc} = \|\mathbf{n} \times \hat{\mathbf{n}} \|
                      + \text{min}\left(\|\mathbf{n} - \hat{\mathbf{n}}\|^2, \|\mathbf{n} + \hat{\mathbf{n}}\|^2 \right).
\end{equation}
Thus, we obtain the loss $\mathcal{L}_{\mathbf{n}}^q$ for query point normal $\mathbf{n}_q$ and the mean loss $\mathcal{L}_{\mathbf{n}}^p$ for neighboring point normals $\mathbf{n}_{i}$.
Moreover, to facilitate the learning of $\tau$ in Eq.~\eqref{eq:output}, we adopt a weight loss based on coplanarity~\cite{zhang2022geometry},
\begin{equation}
  \mathcal{L}_{\tau} = \frac{1}{N} \sum_{i=1}^{N}(\tau_{i} - \hat{\tau}_{i})^2,
  ~~ \hat{\tau}_{i} = \text{exp} \left(- \frac{(p_i \cdot \hat{\mathbf{n}}_q)^2}{\varepsilon^2} \right),
\end{equation}
where $\varepsilon \!=\! \text{max}(0.0025, ~0.3 \sum_{i=1}^{N}(p_i \cdot \hat{\mathbf{n}}_q)^2 / N)$.
In summary, the final training loss function is given by
\begin{equation}  \label{eq:finalloss}
  \mathcal{L} = \lambda_1 \mathcal{L}_{\mathbf{n}}^q + \lambda_2 \mathcal{L}_{\mathbf{n}}^p + \lambda_3 \mathcal{L}_{\tau},
\end{equation}
where the weight factors $\lambda_1\!=\!0.1$, $\lambda_2\!=\!0.4$ and $\lambda_3\!=\!1.0$ are first set empirically, and then fine-tuned according to the experimental results.

\section{Experiments}

\noindent\textbf{Implementation}.
We follow the same experimental setup in~\cite{guerrero2018pcpnet,li2022hsurf} and train our network only on the training set of the PCPNet dataset~\cite{guerrero2018pcpnet}.
In this dataset, each shape corresponds to a clean point cloud and other point clouds generated by five different strategies, including three levels of Gaussian noise with standard deviations of 0.12\% (low), 0.6\% (medium) and 1.2\% (high) of the shape bounding box diagonal and two types of non-uniform sampling (stripe and gradient).
In each training epoch, we randomly sample 1000 query points from each shape as centers and build patches around them.
Each patch is centered on the query point and normalized to a unit sphere.
We set the input patch size $N\!=\!800$,
the size scale set $N_k\!=\!\{N/b^s\}_{s=0}^{L}$, where $L\!=\!2$ (for the number of size scales) and $b\!=\!2$ (for the size of each scale).
The number of kNN points in the per-point feature extraction is $n_k \!=\! 16$.
We train the model for $800$ epochs, and adopt the AdamW optimizer with an initial learning rate of $0.001$, which is decayed to $1/5$ of the latest value at epochs $\{400,600\}$.

\noindent\textbf{Metric}.
We use the normal angle Root Mean Squared Error (RMSE) as the evaluation metric and the Percentage of Good Points (PGP) to count points with qualified normals~\cite{guerrero2018pcpnet,li2022hsurf}.

\noindent\textbf{NestPC dataset}.
In this study, we present a new dataset for 3D point cloud normal estimation evaluation.
The number of datasets in this research field is very limited.
The data in commonly used point cloud normal estimation datasets, such as PCPNet~\cite{guerrero2018pcpnet} and FamousShape~\cite{li2023shsnet}, have relatively simple topological structures.
To comprehensively evaluate the performance of the proposed algorithm, we collect shapes with complex topology and sample point clouds from the provided mesh data.
The ground-truth normals of the point clouds are calculated from the mesh data and used for evaluation.
As shown in Fig.~\ref{fig:NestPC}, we call this new dataset \emph{NestPC}, which will be publicly available along with our source code.

\begin{table*}[t]
  \centering
  \footnotesize
  \setlength{\tabcolsep}{0.8mm}
  \caption{
    Comparison of PGP-$20^\circ$ on the PCPNet and FamousShape datasets under the highest noise.
    The higher the better.
  }
  \vspace{-0.2cm}
  \begin{tabular}{l|ccccccccccc}
    \toprule
    (\%)          & AdaFit~\cite{zhu2021adafit}
                  & GraphFit~\cite{li2022graphfit}
                  & NeAF~\cite{li2023NeAF}
                  & HSurf-Net~\cite{li2022hsurf}
                  & Du \etal~\cite{du2023rethinking}
                  & NGLO~\cite{li2023neural}
                  & SHS-Net~\cite{li2023shsnet}
                  & CMG-Net~\cite{wu2024cmg}
                  & MSECNet~\cite{xiu2023msecnet}
                  & Ours           \\
    \midrule
    PCPNet        & 77.26  & 77.71    & 77.44  & 77.77     & 77.79     & 77.76  & 77.94    & 77.35    & 77.22    & \textbf{78.05}  \\
    FamousShape   & 43.95  & 43.99    & 44.05  & 44.62     & 43.97     & 44.19  & 44.67    & 43.18    & 44.70    & \textbf{44.99}  \\
    \bottomrule
  \end{tabular} 
  \label{tab:PGP}
  \vspace{-0.2cm}
\end{table*}
\begin{table*}[t]
	\centering
	\footnotesize
	\setlength{\tabcolsep}{0.7mm}
	\caption{
		Normal RMSE on sparse point cloud data with 3000 and 5000 points.
	}
	\vspace{-0.2cm}
	\label{table:sparse}
	\begin{tabular}{c|ccccccccccc}
		\toprule
			& AdaFit~\cite{zhu2021adafit}
			& GraphFit~\cite{li2022graphfit}
			& NeAF~\cite{li2023NeAF}
			& HSurf-Net~\cite{li2022hsurf}
			& Du \etal~\cite{du2023rethinking}
			& NGLO~\cite{li2023neural}
			& SHS-Net~\cite{li2023shsnet}
			& CMG-Net~\cite{wu2024cmg}
			& MSECNet~\cite{xiu2023msecnet}
			& GCNO~\cite{xu2023globally}
			& Ours              \\
		\midrule
		3K 	& 30.46  & 29.56    & 28.64  & 27.74     & 27.18    & 26.78  & 27.22   & 26.13   & 27.32    & 27.87  & \textbf{24.50}    \\
		5K 	& 27.05  & 25.47    & 25.10  & 23.93     & 23.30    & 23.07  & 23.55   & 22.40   & 23.64    & 31.54  & \textbf{20.91}    \\
		\bottomrule
	\end{tabular} 
	\vspace{-0.2cm}
\end{table*}
\begin{table*}[t]
    \footnotesize
    \centering
    \setlength{\tabcolsep}{2.5mm}
    \caption{
      Our method can improve the performance of other methods on the PCPNet dataset.
    }
    \vspace{-0.2cm}
    \begin{threeparttable}
      \begin{tabular}{l|ccccccccc}
        \toprule
                    & PCPNet~\cite{guerrero2018pcpnet}
                    & DeepFit~\cite{ben2020deepfit}
                    & NeAF~\cite{li2023NeAF}
                    & HSurf-Net~\cite{li2022hsurf}
                    & NGLO~\cite{li2023neural}
                    & CMG-Net~\cite{wu2024cmg}
                    \\
        \midrule
        Original      & 14.58                    & 11.80                     & 10.22                    & 10.11                   & 9.98                    & 9.93                      \\   
        w/ Ours-1     & 9.71  ($\downarrow$4.87) & 9.69  ($\downarrow$2.11)  & 9.69  ($\downarrow$0.53) & 9.69 ($\downarrow$0.42) & 9.69 ($\downarrow$0.29) & 9.69 ($\downarrow$0.24)   \\   
        w/ Ours-2     & 10.30 ($\downarrow$4.28) & 10.75 ($\downarrow$1.05)  & 10.06 ($\downarrow$0.16) & 9.82 ($\downarrow$0.29) & 9.71 ($\downarrow$0.27) & 9.87 ($\downarrow$0.06)   \\   
        \bottomrule
      \end{tabular}  
      \begin{tablenotes}
        \footnotesize
        \item \noindent `Ours-1' indicates normal optimization and `Ours-2' indicates network module integration.
      \end{tablenotes}
    \end{threeparttable}
    \label{tab:extend}
    \vspace{-0.2cm}
\end{table*}
\begin{table*}[t]
    \footnotesize
    \centering
    \setlength{\tabcolsep}{.8mm}
    \caption{
      Comparison of the network parameter (million) and the average inference time (seconds per 100k points).
    }
    \vspace{-0.2cm}
    \begin{threeparttable}
      \begin{tabular}{l|cccccccccc}
        \toprule
                      & AdaFit~\cite{zhu2021adafit}
                      & GraphFit~\cite{li2022graphfit}
                      & NeAF~\cite{li2023NeAF}
                      & HSurf-Net~\cite{li2022hsurf}
                      & Du \etal~\cite{du2023rethinking}
                      & NGLO~\cite{li2023neural}
                      & SHS-Net~\cite{li2023shsnet}
                      & CMG-Net~\cite{wu2024cmg}
                      & MSECNet~\cite{xiu2023msecnet}
                      & Ours                      \\
        \midrule
        Param.        & 4.87   & 4.26     & 6.74   & 2.16      & 4.46     & 0.46+1.92  & 3.27    & 2.70    & 10.40       & \textbf{2.03}             \\
        Time          & 56.23  & 292.12   & 400.81 & 72.47     & 295.69   & 0.56+70.77 & 65.89   & 109.98  & 0.61$^*$    & \textbf{5.06 (0.11$^*$)}  \\
        \bottomrule
      \end{tabular} 
      \begin{tablenotes}
        \footnotesize
        \item $*$ represents patch-to-patch normal estimation for the entire shape~\cite{xiu2023msecnet}.
      \end{tablenotes}
    \end{threeparttable}
    \label{tab:time}
    \vspace{-0.2cm}
\end{table*}

\begin{table}[t]
\centering
\footnotesize
\setlength{\tabcolsep}{.7mm}
\caption{
    Normal RMSE on the NestPC dataset.
}
\label{table:nestpc}
\vspace{-0.2cm}
\resizebox{\linewidth}{!}{
\begin{tabular}{cccccccc}
    \toprule
    GraphFit     & HSurf-Net  & Du \etal   & NGLO    & SHS-Net  & CMG-Net   & MSECNet  & Ours             \\
    \midrule
    18.44        & 13.38      & 13.36      & 12.79   & 13.37    & 13.99     & 13.89    & \textbf{11.62}   \\
    \bottomrule
\end{tabular} }
\vspace{-0.2cm}
\end{table}

\begin{table}[t]
\centering
\footnotesize
\setlength{\tabcolsep}{.75mm}
\caption{
	Normal RMSE on the SceneNN dataset.
}
\label{table:scenenn}
\vspace{-0.2cm}
\begin{tabular}{c|ccccccc}
	\toprule
			& HSurf-Net
			& Du \etal
			& NGLO
			& SHS-Net
			& CMG-Net
			& MSECNet
			& Ours   \\
	\midrule
	Clean       & 7.55   & 7.68       & 7.73   & 7.93     & 7.64      & \textbf{6.94}     & 7.28   \\
	Noise       & 12.23  & 11.72      & 12.25  & 12.40    & 11.82     & 11.66             & \textbf{11.47}  \\
	\bottomrule
\end{tabular} 
\vspace{-0.2cm}
\end{table}

\subsection{Baseline Methods}

In our experiments, we compare our method with various baseline methods, which are mainly classified into the following three categories:
(1) the traditional normal estimation methods, such as PCA~\cite{hoppe1992surface}, Jet~\cite{cazals2005estimating} and GCNO~\cite{xu2023globally};
(2) the learning-based methods using the surface fitting, such as DeepFit~\cite{ben2020deepfit}, AdaFit~\cite{zhu2021adafit}, GraphFit~\cite{li2022graphfit} and Du \etal~\cite{du2023rethinking};
(3) the learning-based methods for normal regression, such as PCPNet~\cite{guerrero2018pcpnet}, Nesti-Net~\cite{ben2019nesti}, Refine-Net~\cite{zhou2022refine}, HSurf-Net~\cite{li2022hsurf}, SHS-Net~\cite{li2023shsnet,li2024shsnet-pami}, NeuralGF~\cite{li2023neuralgf} and MSECNet~\cite{xiu2023msecnet}.

For quantitative comparisons on the PCPNet dataset~\cite{guerrero2018pcpnet}, the numerical results of some baseline methods are taken from their papers due to the unavailable source codes, including Zhou \etal~\cite{zhou2020normal}, MTRNet~\cite{cao2021latent} and Zhou \etal~\cite{zhou2023improvement}.
For the method of Zhang \etal~\cite{zhang2022geometry}, we only feed the 3D point clouds into their network model to predict normals since their precomputed features are not available.
For GraphFit~\cite{li2022graphfit}, we train a neural network model from scratch using the official source code.
For the algorithm of Du \etal~\cite{du2023rethinking}, we use GraphFit~\cite{li2022graphfit} as the backbone network.

\subsection{Results on Shape Data}

The quantitative comparison results on the datasets PCPNet~\cite{guerrero2018pcpnet} and FamousShape~\cite{li2023shsnet} are reported in Table~\ref{table:pcpnet_famousShape}.
Our method achieves significant performance improvement, and provides more accurate normal results than baseline methods under most noise levels and density variations.
We show the normal PGP of different data categories in Fig.~\ref{fig:PGP_pcpnet_4} and Table~\ref{tab:PGP}.
We can see that our method has the best performance at almost all thresholds.
In Fig.~\ref{fig:PGP_pcpnet_2}, we show more results of normal PGP on noisy point clouds of the PCPNet dataset.
It can be seen that our method has a performance advantage compared to the baseline methods.
Fig.~\ref{fig:error_pcpnet} shows a visual comparison of the normal errors on different shapes, where the point clouds are rendered in RGB colors according to the error map.
We provide the quantitative comparison results on our NestPC dataset in Table~\ref{table:nestpc}.
Moreover, Fig.~\ref{fig:errorMap} shows a visual comparison of normal errors on two shapes with complex topology and geometry.
These evaluation results all demonstrate the excellent performance of our algorithm.

\noindent\textbf{Sparse data}.
We perform a quantitative evaluation on two sets of point clouds that have the same shapes as the FamousShape dataset~\cite{li2023shsnet}, but each shape in these two sets contains only 3000 and 5000 points, respectively.
As shown in Table~\ref{table:sparse}, we report quantitative comparison results of unoriented normals on these two data sets.
We can see that our method has the lowest RMSE result.
Fig.~\ref{fig:PGP_sparse} shows the normal PGP of each evaluated method.
These results demonstrate the good performance of our method on sparse point clouds.

\begin{figure}[t]
  \centering
  \includegraphics[width=\linewidth]{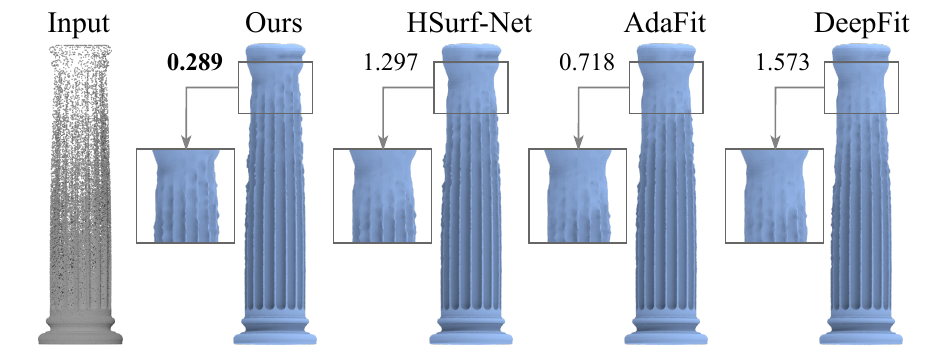}  \vspace{-0.7cm}
  \caption{
    Comparison of the reconstructed surfaces from a non-uniformly sampled point cloud using the estimated normals.
    The chamfer distance ($\times 10^{-4}$) is provided.
  }
  \label{fig:reconstruction}
  \vspace{-0.2cm}
\end{figure}

\begin{figure}[t]
  \centering
  \includegraphics[width=\linewidth]{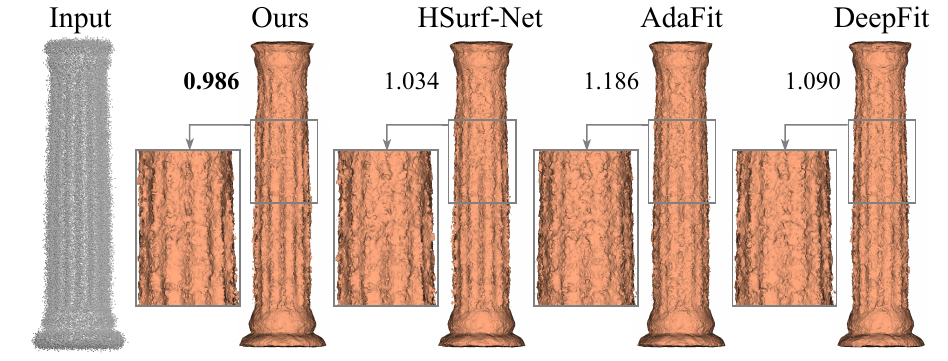}  \vspace{-0.7cm}
  \caption{
    Comparison of the reconstructed surfaces from the denoised point clouds.
    The noisy point cloud is first denoised using the estimated normals.
    The chamfer distance ($\times 10^{-4}$) is provided.
  }
  \label{fig:denoising}
  \vspace{-0.2cm}
\end{figure}

\begin{figure*}[t]
	\centering
	\includegraphics[width=\linewidth]{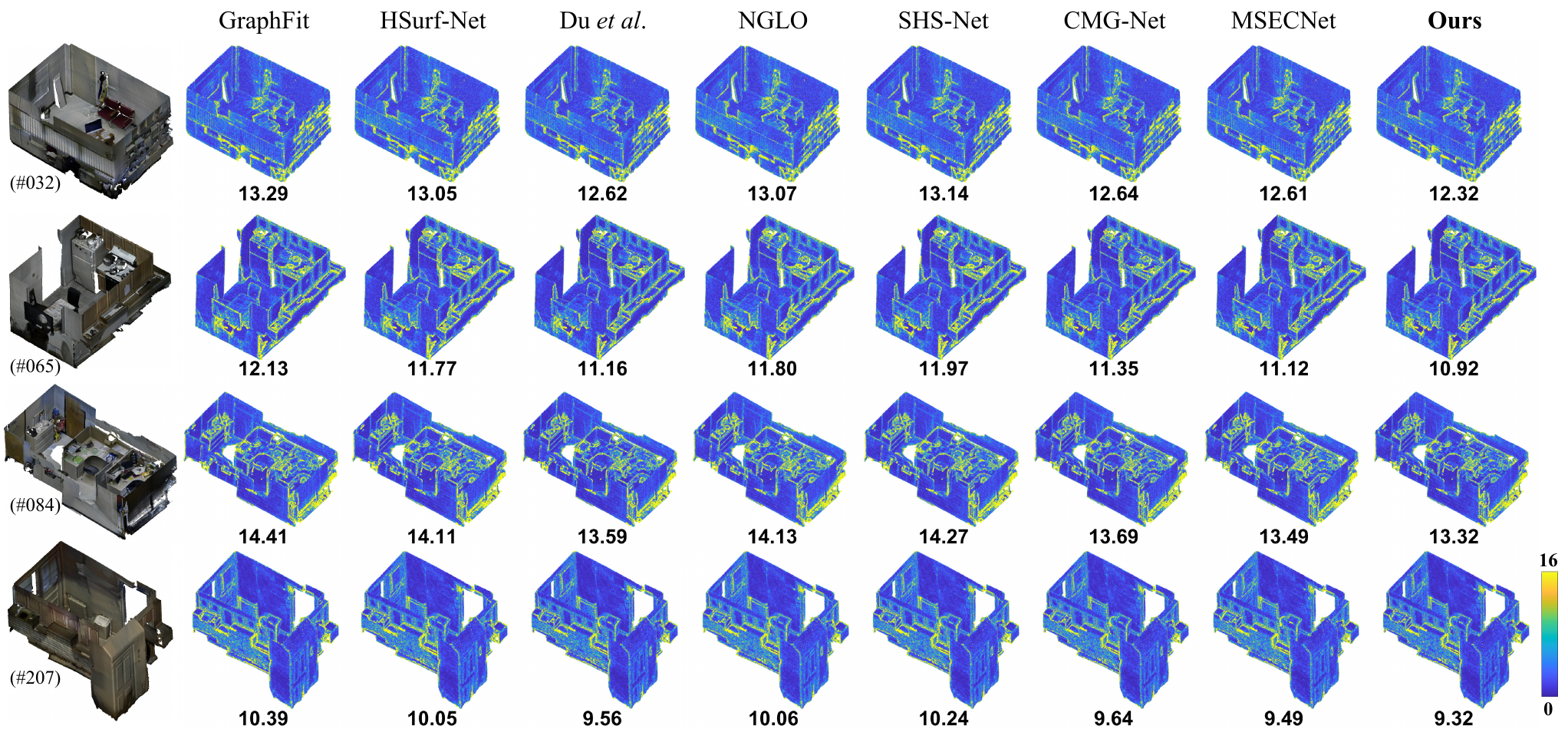}  \vspace{-0.7cm}
  \caption{
    Comparison of normal RMSE on noisy point clouds of the SceneNN dataset.
    The colors on point clouds denote the normal errors mapped to a heatmap ranging from $0^{\circ}$ to $16^{\circ}$.
    The average RMSE of the entire point cloud is provided under each point cloud.
    The first column shows real-world indoor scenes with RGB colors.
  }
  \label{fig:errorMap_scenenn}
\end{figure*}

\begin{figure*}[t]
  \centering
  \includegraphics[width=\linewidth]{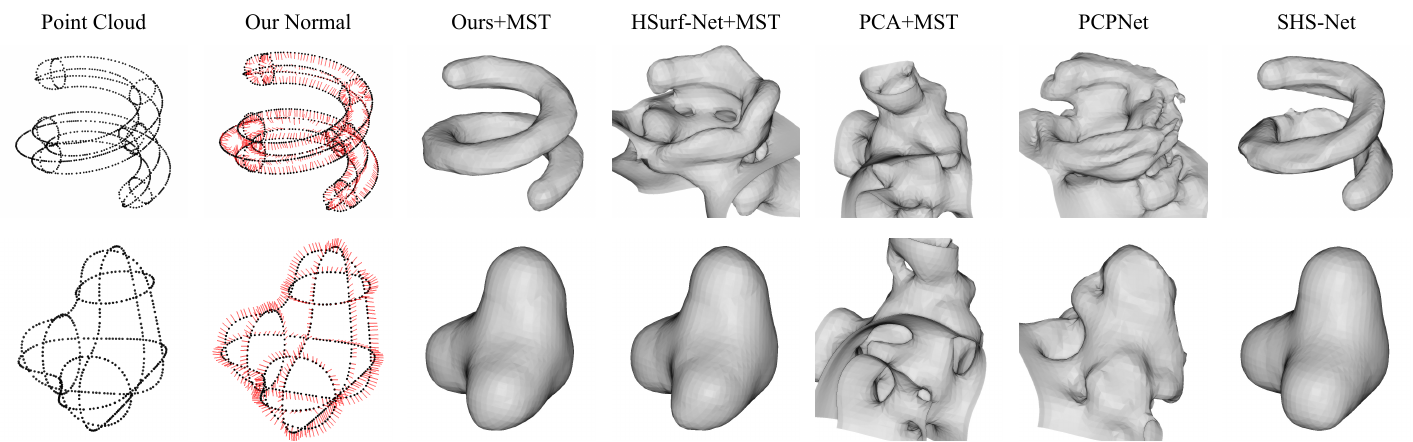}  \vspace{-0.6cm}
  \caption{
    Visual comparison on wireframe point clouds with sparse and non-uniform sampling.
    The input point clouds contain less than 1000 points.
    Our normals (red lines), as well as the normals of HSurf-Net and PCA, are reoriented via MST~\cite{hoppe1992surface}, and PCPNet and SHS-Net can estimate oriented normals.
  }
  \label{fig:wireframe}
\end{figure*}

\begin{figure*}[t]
  \centering
  \includegraphics[width=\linewidth]{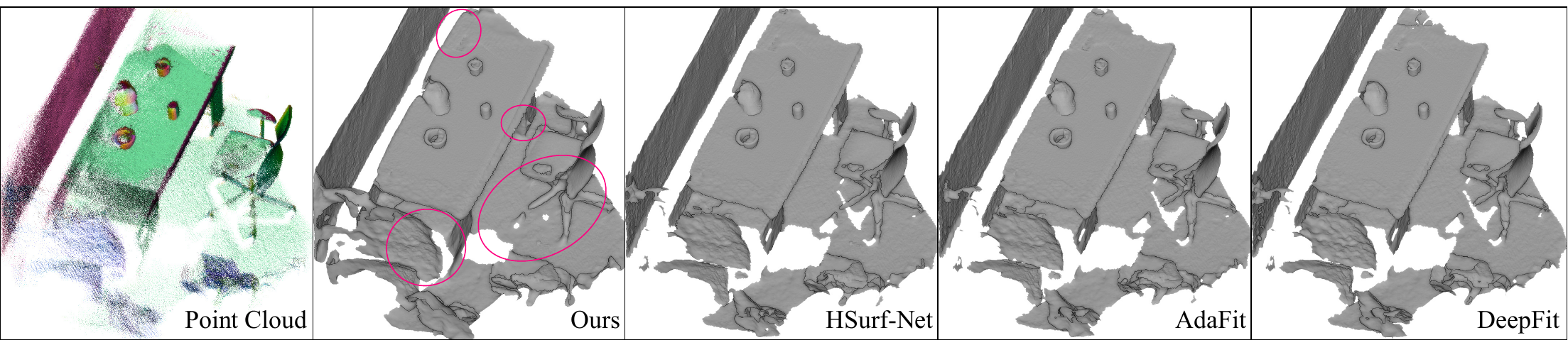}  \vspace{-0.7cm}
  \caption{
    Surfaces reconstructed using normals estimated by different methods on the RGB-D Scenes Dataset.
    The generated surfaces are cropped using the raw point cloud, leaving holes in the wrong surfaces.
  }
  \label{fig:poissonRecon_Others-14}
\end{figure*}

\begin{figure*}[t]
  \centering
  \includegraphics[width=\linewidth]{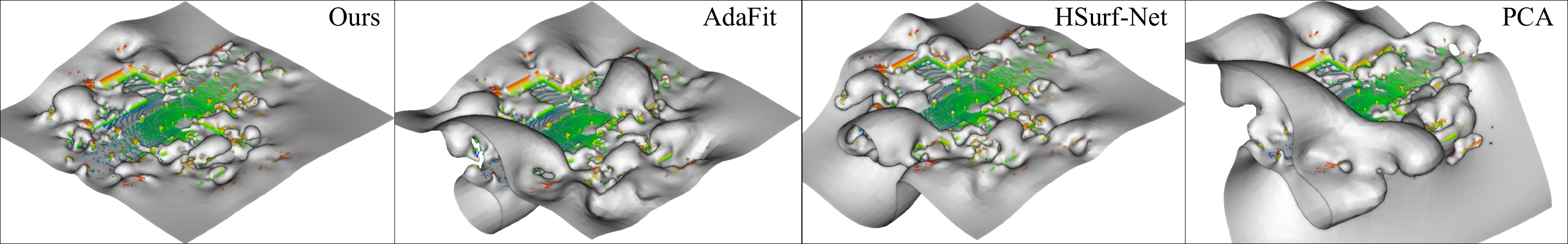}  \vspace{-0.7cm}
  \caption{
    The reconstructed surfaces using normals estimated by different methods on the KITTI dataset.
    Points are colored by their height values.
  }
  \label{fig:kitti_mesh}
  \vspace{-0.2cm}
\end{figure*}

\begin{figure*}[t]
  \centering
  \includegraphics[width=.75\linewidth]{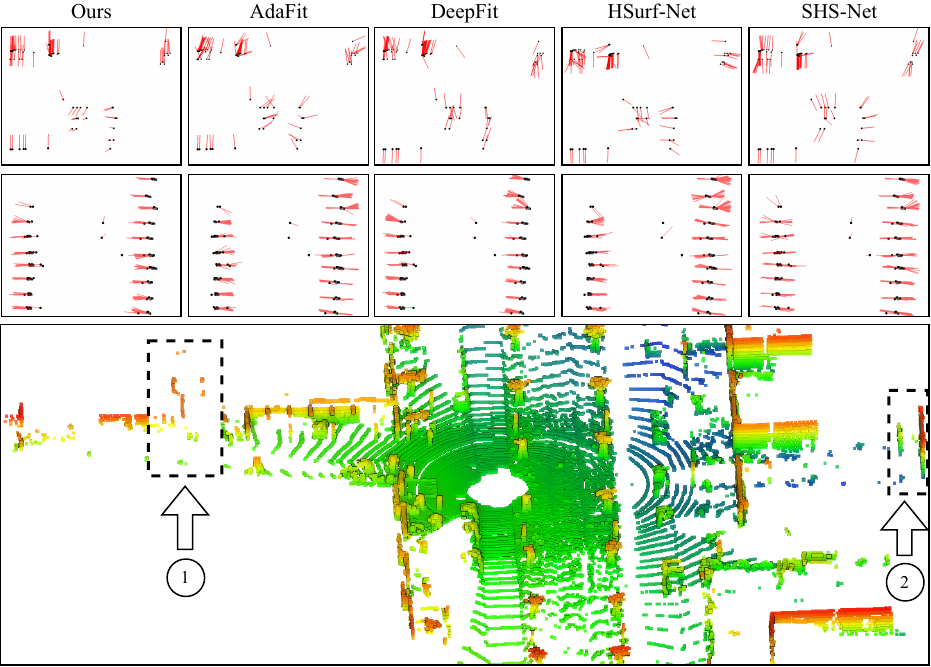}  \vspace{-0.3cm}
  \caption{
    Visual comparison of the estimated normal vectors (red lines in the first two rows) on the KITTI dataset.
    In extremely sparse and noisy planar regions, the parallelism between our estimated normals is better.
    Points in the third row are colored by their height values.
  }
  \label{fig:kitti_normal}
\end{figure*}

\begin{figure*}[t]
  \centering
  \subfloat[]{
    \includegraphics[width=.3\linewidth]{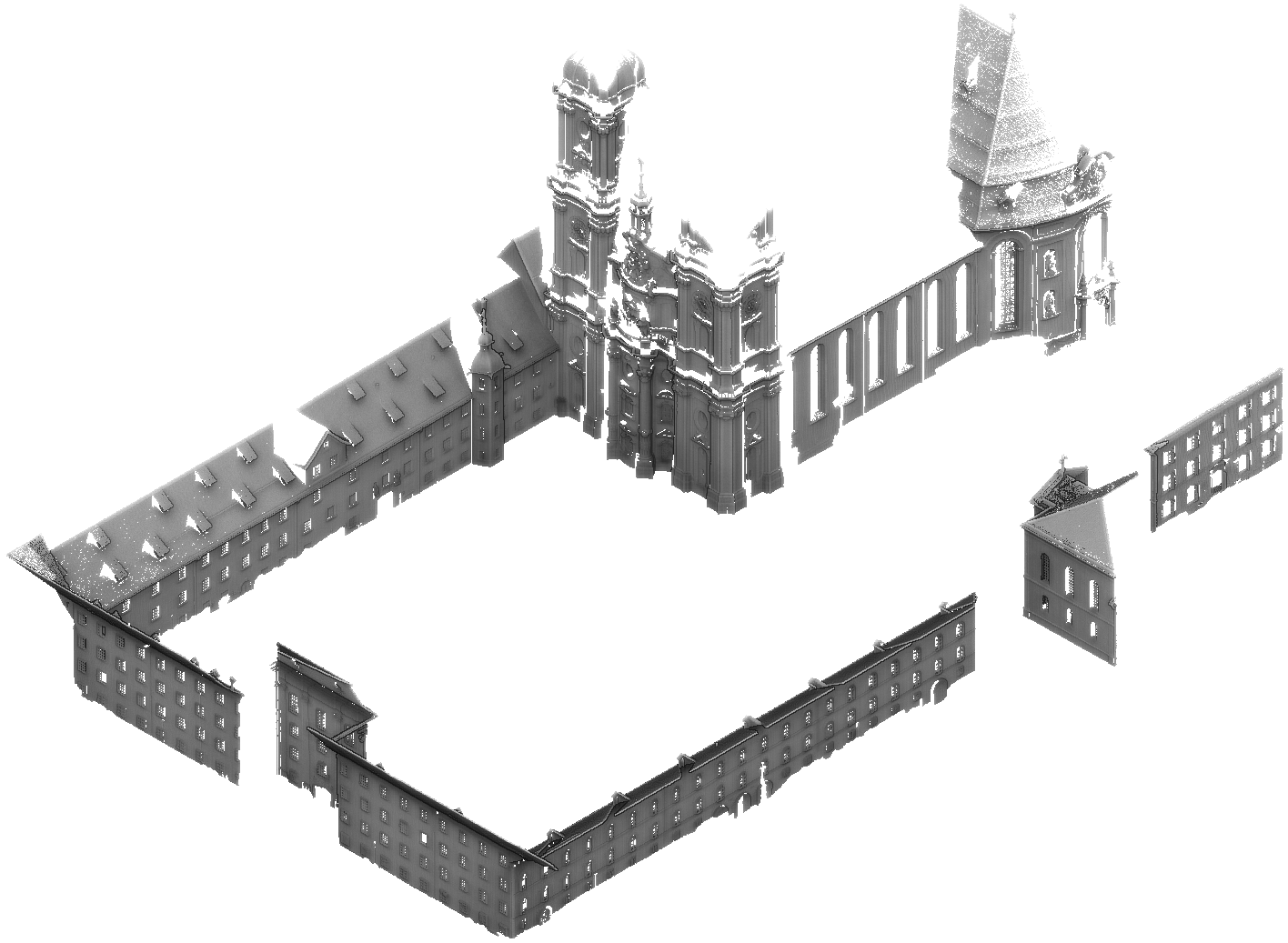}
  }
  \quad
  \subfloat[]{
    \includegraphics[width=.3\linewidth]{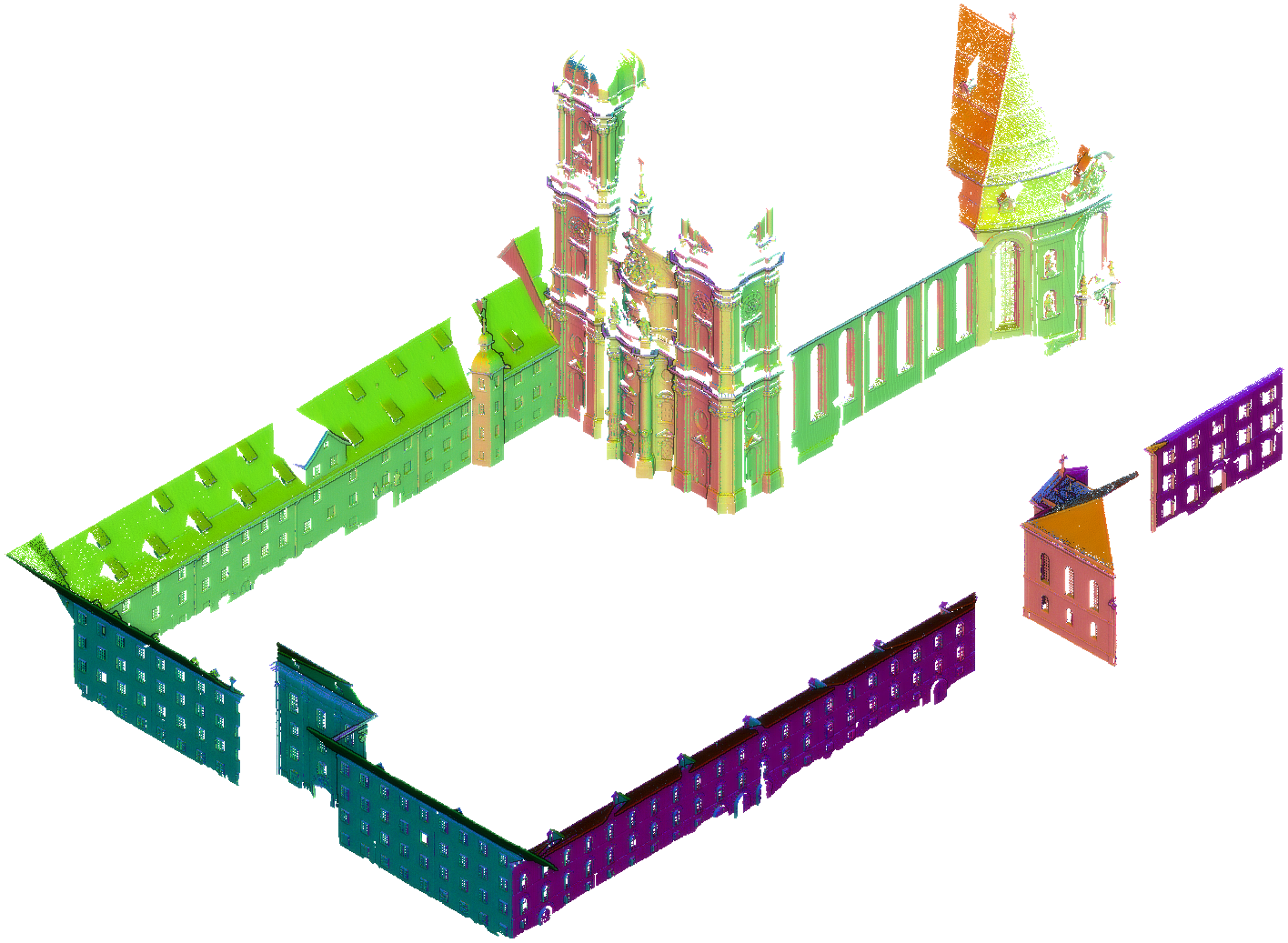}
  }
  \quad
  \subfloat[]{
    \includegraphics[width=.3\linewidth]{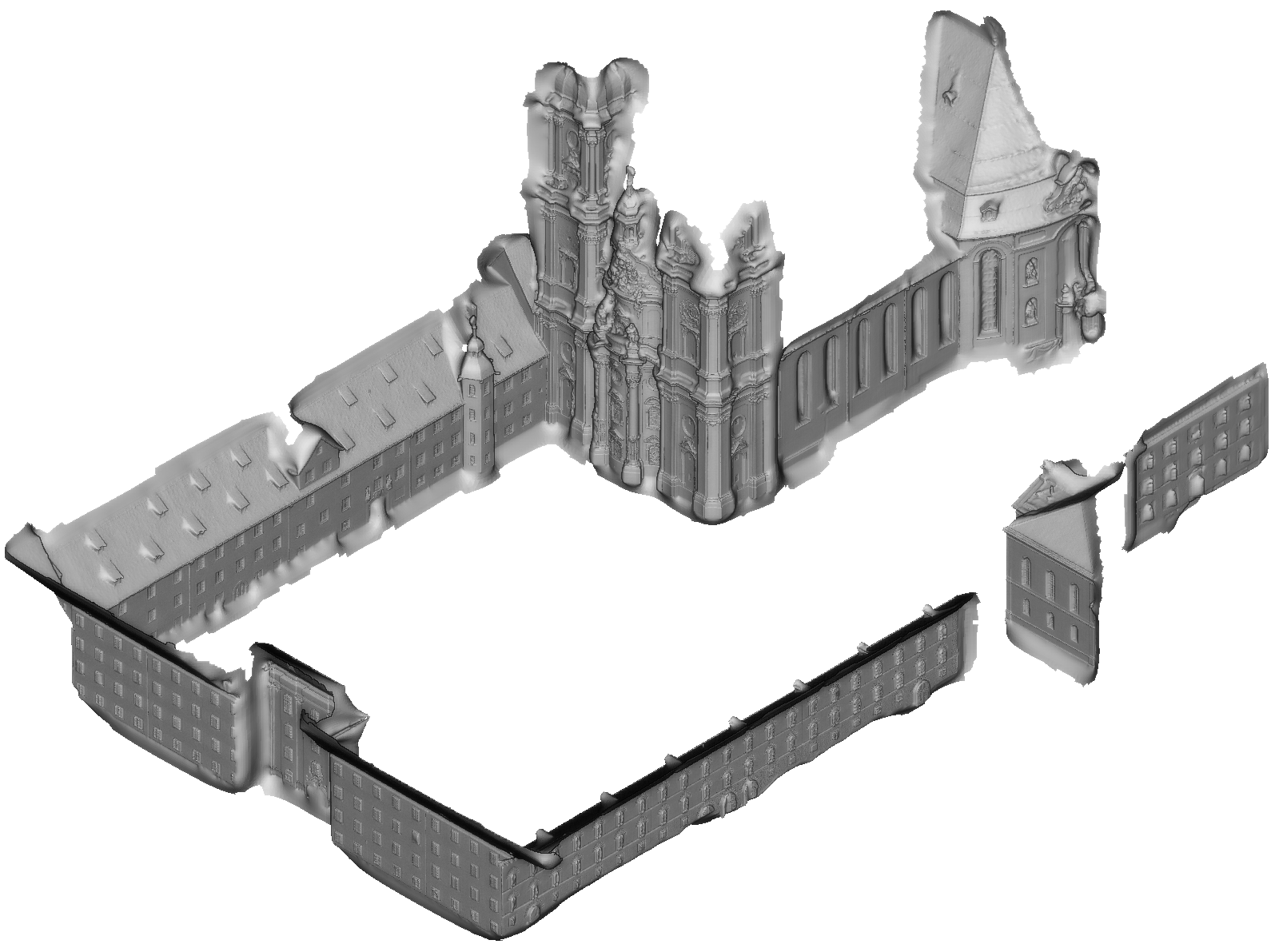}
  }
  \vspace{-0.2cm}
  \caption{
    Visualization of our estimated normals and reconstructed surface on a point cloud of the Semantic3D dataset.
    (a) The input LiDAR point cloud of the real-world outdoor scene.
    (b) The normal estimation results of our method.
    The estimated normal vectors are mapped to RGB colors for visualization, where different colors represent different normal directions~\cite{ben2019nesti}.
    (c) The reconstructed surface using estimated normals.
  }
  \label{fig:Semantic3D}
  \vspace{-0.2cm}
\end{figure*}

\begin{figure}[t]
	\centering
	\includegraphics[width=\linewidth]{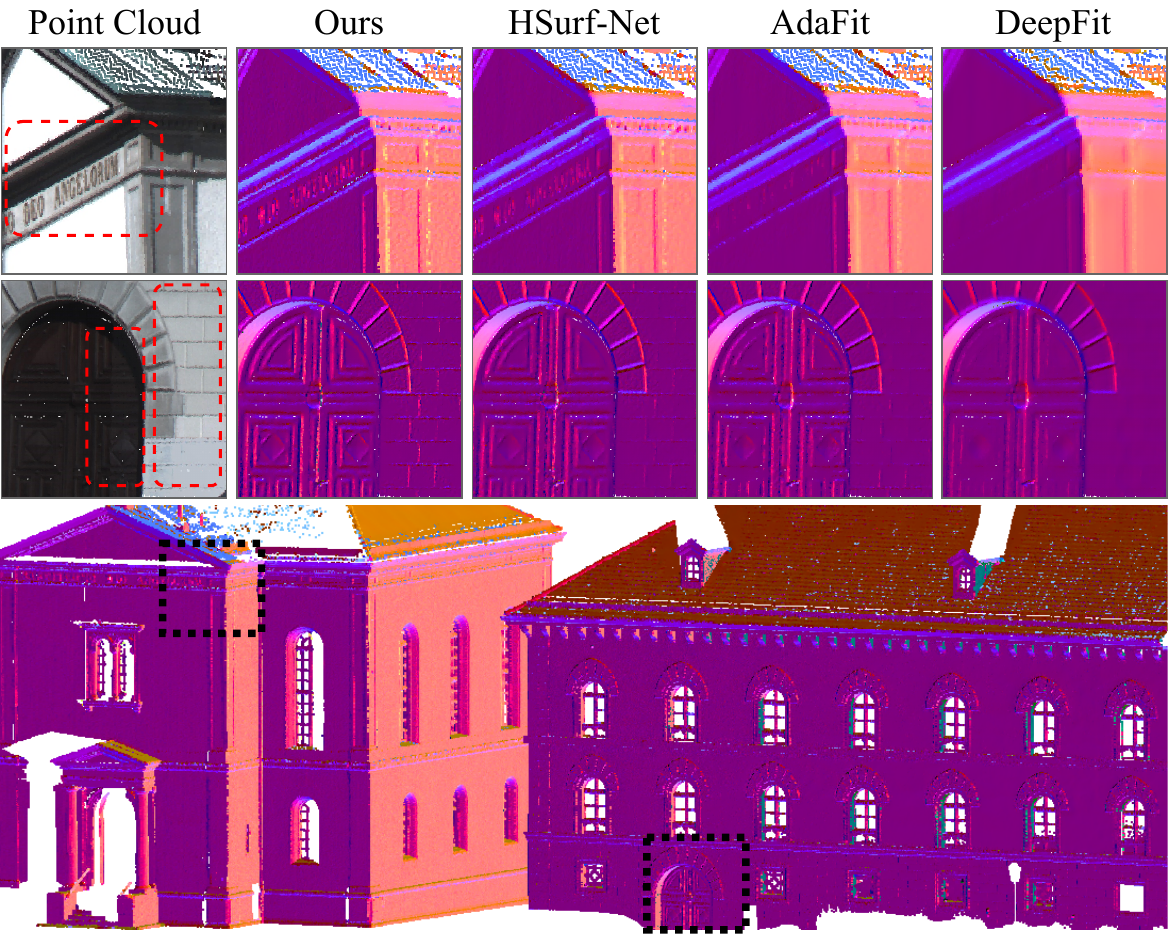}  \vspace{-0.7cm}
  \caption{
  Visual comparison of normal results on the Semantic3D dataset.
  We compare the local details of the building in the first two rows and show our estimated normals of the building in the third row.
  The point normal vectors are mapped to RGB colors.
  }
  \label{fig:map_semantic3d}
\end{figure}

\subsection{Results on Scene Data}

To evaluate the generalization ability of our method, we directly use the model trained on the PCPNet shape dataset to do testing on the real scene data.
The SceneNN dataset~\cite{hua2016scenenn} provides RGB-D data captured in various real-world room scenes and all scenes are reconstructed as triangle meshes.
We use the preprocessed data from \cite{li2022hsurf} for normal evaluation.
The ground-truth normals of this dataset are calculated from the provided mesh data.
We report the quantitative comparison results in Table~\ref{table:scenenn}, and our method achieves better performance than baseline methods in the data category of noise.
In Fig.~\ref{fig:errorMap_scenenn}, we visualize the normal RMSE on point clouds of several indoor scenes provided by the SceneNN dataset.
Our method achieves the lowest normal estimation error in these scenes with complex geometries.

\subsection{Applications}

\noindent\textbf{Improve other methods}.
We provide two types of experiments to illustrate that our algorithm can further improve the performance of other methods:

(1) \emph{Ours-1}: We do not make any modifications or adjustments to other methods, but directly optimize their output results.
Following the normal optimization in~\cite{li2023neural}, we first modify our network and train it to predict the angle distance between an arbitrary input vector and the true normal, rather than predicting the normal.
Then, we construct vector samples in spherical space based on the normal results of other methods.
Finally, the vector samples are used as the input to predict the angle of each sample, and the sample with the smallest output angle is selected as the optimal normal.

(2) \emph{Ours-2}: We replace the main structure of the feature extraction network of other methods with our designed modules, namely multi-scale feature aggregation layer and cross-scale compensation layer.
We do not change the input data and training losses of other methods, and keep the parameters of the new network models as close as possible to the same order of magnitude as their original network models.
The results of the above two experiments (\ie, Ours-1 and Ours-2) are shown in Table~\ref{tab:extend} and Fig.~\ref{fig:extension}, which demonstrate the good scalability and portability of our approach.

\noindent
\textbf{Surface reconstruction}.
To investigate the effect of the estimated normals, we use Poisson reconstruction~\cite{kazhdan2006poisson} to reconstruct object or scene surfaces from point clouds.
The reconstructed object surfaces using the estimated normals are visualized in Fig.~\ref{fig:reconstruction}.
More surface reconstruction results of scene data are shown in Figures~\ref{fig:poissonRecon_Others-14} and \ref{fig:kitti_mesh}.
The results show that the reconstruction algorithm can benefit from the normals estimated by our method to recover more detail in sparse or sharp areas.
We also provide qualitative comparison results on wireframe-type point clouds with less than 1000 points.
As shown in Fig.~\ref{fig:wireframe}, our method can handle sparse and non-uniformly sampled point clouds.

\noindent
\textbf{Point cloud denoising}.
We adopt a denoising method~\cite{lu2020low} with default parameters to filter the input noisy point cloud based on the normals estimated by different methods.
The comparison of reconstructed surfaces from the denoised point clouds is shown in Fig.~\ref{fig:denoising}.
The results indicate that our estimated normals can facilitate the denoising algorithm to keep more complete structures and details.

\subsection{More Results on Different Data}

In this section, we will show that our method (only trained on the PCPNet shape data) can generalize well to new and different types of point cloud data, such as RGB-D data of indoor scenes and LiDAR data of outdoor scenes.

\noindent
\textbf{Kinect data}.
We use the RGB-D Scenes Dataset~\cite{lai2014unsupervised} to test the generalization ability of our method to real-world scene data.
This dataset is captured in indoor room scenes and the ground-truth normal is unavailable.
We adopt ODP algorithm~\cite{metzer2021orienting} to make the estimated normals have a consistent orientation and employ Poisson reconstruction algorithm~\cite{kazhdan2013screened} to reconstruct surfaces.
As shown in Fig.~\ref{fig:poissonRecon_Others-14}, we visualize the surface reconstruction results based on the normals estimated by different methods.
The results show that our method gives better shapes of objects in the room.

\noindent
\textbf{KITTI dataset}.
The KITTI dataset~\cite{geiger2012we} is captured from real-world street scenes using a laser scanner, but its point cloud data is sparser in points compared to the Paris-rue-Madame dataset.
Since the ground-truth normal is unavailable, we use ODP algorithm~\cite{metzer2021orienting} and Poisson reconstruction algorithm~\cite{kazhdan2013screened} to reconstruct surfaces from the estimated normals.
As shown in Fig.~\ref{fig:kitti_normal}, we show a visual comparison of the estimated normals.
In Fig.~\ref{fig:kitti_mesh}, we provide a visual comparison of the reconstructed surfaces based on the normals.

\noindent
\textbf{Semantic3D dataset}.
Different from the above datasets, the Semantic3D dataset~\cite{hackel2017isprs} is obtained by scanning different scenic spots or buildings with LiDAR.
It has a larger scene scale and contains more points in a single point cloud.
As the ground truth normal of the Semantic3D dataset is not available, we show a visual comparison in Fig.~\ref{fig:map_semantic3d}.
We can see that our method reveals the fine details of buildings, while the baseline methods perform over-smooth in these complex structure areas.
In Fig.~\ref{fig:Semantic3D}, we show our estimated point cloud normals on a large-scale outdoor scene of the Semantic3D dataset.
Since the ground-truth normals are not available, we map the estimated normal vectors to RGB colors for visualization and reconstruct the corresponding surface.

The above evaluation results demonstrate that our model trained with local patches on shape data can generalize well to real-world LiDAR data.

\begin{table*}[t]
\centering
\footnotesize
\setlength{\tabcolsep}{2.5mm}
\caption{
    Ablation studies on the PCPNet dataset.
    The discussion is provided in the text.
}
\vspace{-0.2cm}
\label{tab:ablation}
\begin{tabular}{ll|cccc||cccc|cc| >{\columncolor{mygray}} c}
    \toprule
    \multicolumn{2}{c|}{\multirow{2}{*}{\textbf{Ablation}}} & \multirow{2}{*}{\shortstack{Per-point\\Feature}}
    & \multirow{2}{*}{\shortstack{Block \\ $\mathcal{F}_1$}} & \multirow{2}{*}{\shortstack{Block \\ $\mathcal{F}_2$}} & \multirow{2}{*}{\shortstack{Attention}}
    & \multicolumn{4}{c|}{Noise} & \multicolumn{2}{c|}{{Density}} &   \\
    & & & & & & None & Low & Medium & High     & Stripe & Gradient                     & \multirow{-2}{*}{{Average}} \\
    \midrule
    \multirow{2}{*}{\textbf{(a)}}
    & PointNet-like & & \checkmark & \checkmark & \checkmark & 4.63 & 8.91 & 16.11 & 21.09 & 5.31 & 5.14 &   10.20 \\
    & DGCNN-like    & & \checkmark & \checkmark & \checkmark & 3.83 & 8.43 & 15.78 & 20.96 & 4.62 & 4.36 &   9.66  \\
    \hline
    \multirow{3}{*}{\textbf{(b)}}
    & w/o weight $w$             & \checkmark & & \checkmark & \checkmark & 3.71 & 8.50 & 16.05 & 21.46 & 4.61 & 4.36 &   9.78  \\
    & w/ other weight            & \checkmark & & \checkmark & \checkmark & 3.61 & 8.46 & 16.01 & 21.47 & 4.44 & 4.28 &   9.71  \\
    & w/o $\mathcal{F}_1$        & \checkmark & & \checkmark & \checkmark & 4.18 & 8.69 & 16.53 & 21.93 & 5.13 & 4.91 &   10.23 \\
    \hline
    \multirow{2}{*}{\textbf{(c)}}
    & w/o $\mathcal{F}_2$                 & \checkmark & \checkmark & & \checkmark & 3.69 & 8.50 & 16.05 & 21.41 & 4.67 & 4.55 &   9.81 \\
    & $\mathcal{F}_2 \to \mathcal{F}_1$   & \checkmark & \checkmark & & \checkmark & 3.59 & 8.48 & 15.78 & 21.03 & 4.64 & 4.23 &   9.62 \\
    \hline
    \multirow{4}{*}{\textbf{(d)}}
    & w/ \emph{softmax}-1  & \checkmark & \checkmark & \checkmark & & 3.39 & 8.41 & 15.62 & 20.97 & \textbf{4.01} & 3.96 &   9.39 \\
    & w/ \emph{softmax}-2  & \checkmark & \checkmark & \checkmark & & 3.41 & 8.34 & 15.68 & 20.97 & 4.22 & 4.06 &   9.45 \\
    & w/ simp. concat      & \checkmark & \checkmark & \checkmark & & 3.73 & 8.33 & 15.78 & 21.01 & 4.57 & 4.36 &   9.63 \\
    & w/ simp. add         & \checkmark & \checkmark & \checkmark & & 3.57 & 8.35 & 15.69 & 20.99 & 4.20 & 4.03 &   9.47 \\
    & w/ add               & \checkmark & \checkmark & \checkmark & & 3.33 & 8.36 & 15.70 & 20.95 & 4.11 & 4.02 &   9.41 \\
    \hline
    \multirow{3}{*}{\textbf{(e)}}
    & w/o $d_{\rm sin}$                  & \checkmark & \checkmark & \checkmark & \checkmark & 3.73 & 8.45 & \textbf{15.60} & \textbf{20.82} & 4.49 & 4.25 &   9.56   \\
    & w/o $d_{\rm euc}$                  & \checkmark & \checkmark & \checkmark & \checkmark & 3.66 & 8.37 & 16.05 & 21.33 & 4.38 & 4.31 &   9.68   \\
    & w/o $\mathcal{L}_{\mathbf{n}}^p$   & \checkmark & \checkmark & \checkmark & \checkmark & 3.67 & 8.36 & 15.76 & 21.05 & 4.51 & 4.31 &   9.61   \\
    & w/o $\mathcal{L}_{\tau}$           & \checkmark & \checkmark & \checkmark & \checkmark & 4.41 & 8.68 & 15.77 & 21.00 & 5.41 & 5.16 &   10.07  \\  
    \hline
    \multirow{5}{*}{\textbf{(f)}}
    & $N\!=\!500$  & \checkmark & \checkmark & \checkmark & \checkmark & 3.33 & \textbf{8.26} & 15.74 & 21.39 & 4.17 & 4.09 &   9.49  \\
    & $N\!=\!600$  & \checkmark & \checkmark & \checkmark & \checkmark & 3.39 & 8.31 & 15.66 & 21.11 & 4.17 & 4.02 &   9.44  \\
    & $N\!=\!700$  & \checkmark & \checkmark & \checkmark & \checkmark & 3.43 & 8.29 & 15.68 & 21.02 & 4.13 & 4.03 &   9.43   \\
    & $N\!=\!900$  & \checkmark & \checkmark & \checkmark & \checkmark & 3.45 & 8.29 & 15.69 & 20.94 & 4.39 & \textbf{3.92} &   9.45   \\
    & $N\!=\!1000$ & \checkmark & \checkmark & \checkmark & \checkmark & 3.38 & 8.37 & 15.64 & 20.87 & 4.18 & 4.03 &   9.41  \\
    \midrule
    & Ours             & \checkmark & \checkmark & \checkmark & \checkmark & \textbf{3.32} & {8.34}  & {15.63} & {20.94} & {4.10} & \textbf{3.92} &   \textbf{9.38}  \\
    \bottomrule
\end{tabular} 
\vspace{-0.3cm}
\end{table*}

\begin{table}[t]
\centering
\footnotesize
\setlength{\tabcolsep}{1.0mm}
\caption{
  Normal RMSE on the PCPNet dataset using various patch sizes.
  }
\vspace{-0.2cm}
\label{table:pcpnet_1}
\begin{tabular}{ccccccc}
  \toprule
  AdaFit  & HSurf-Net  & CMG-Net   & MSECNet  & \tabincell{c}{Ours \\(N=500)} & \tabincell{c}{Ours \\(N=700)}  & \tabincell{c}{Ours \\(N=1000)}  \\
  \midrule
  10.76   & 10.11      & 9.93      & 9.76     & 9.49                          & 9.43                           & \textbf{9.41}                            \\
  \bottomrule
\end{tabular} 
\vspace{-0.2cm}
\end{table}

\begin{table}[t]
  \centering
  \footnotesize
  \setlength{\tabcolsep}{.9mm}
  \caption{
    Normal RMSE on the PCPNet dataset using various patch sizes.
  }
  \vspace{-0.2cm}
  \label{table:pcpnet_2}
  \begin{tabular}{c|ccccc}
    \toprule
    Size & AdaFit~\cite{zhu2021adafit} & HSurf-Net~\cite{li2022hsurf} & CMG-Net~\cite{wu2024cmg} & MSECNet~\cite{xiu2023msecnet} & Ours \\
    \midrule
    800  & 10.85  & 10.13     & 9.93    & 9.78    & \textbf{9.38} \\
    900  & 10.89  & 10.23     & 9.99    & 9.79    & \textbf{9.36} \\
    1000 & 10.96  & 10.39     & 10.04   & 9.80    & \textbf{9.36} \\
    \bottomrule
  \end{tabular} 
  \vspace{-0.2cm}
\end{table}

\begin{figure}[t]
  \centering
  \includegraphics[width=\linewidth]{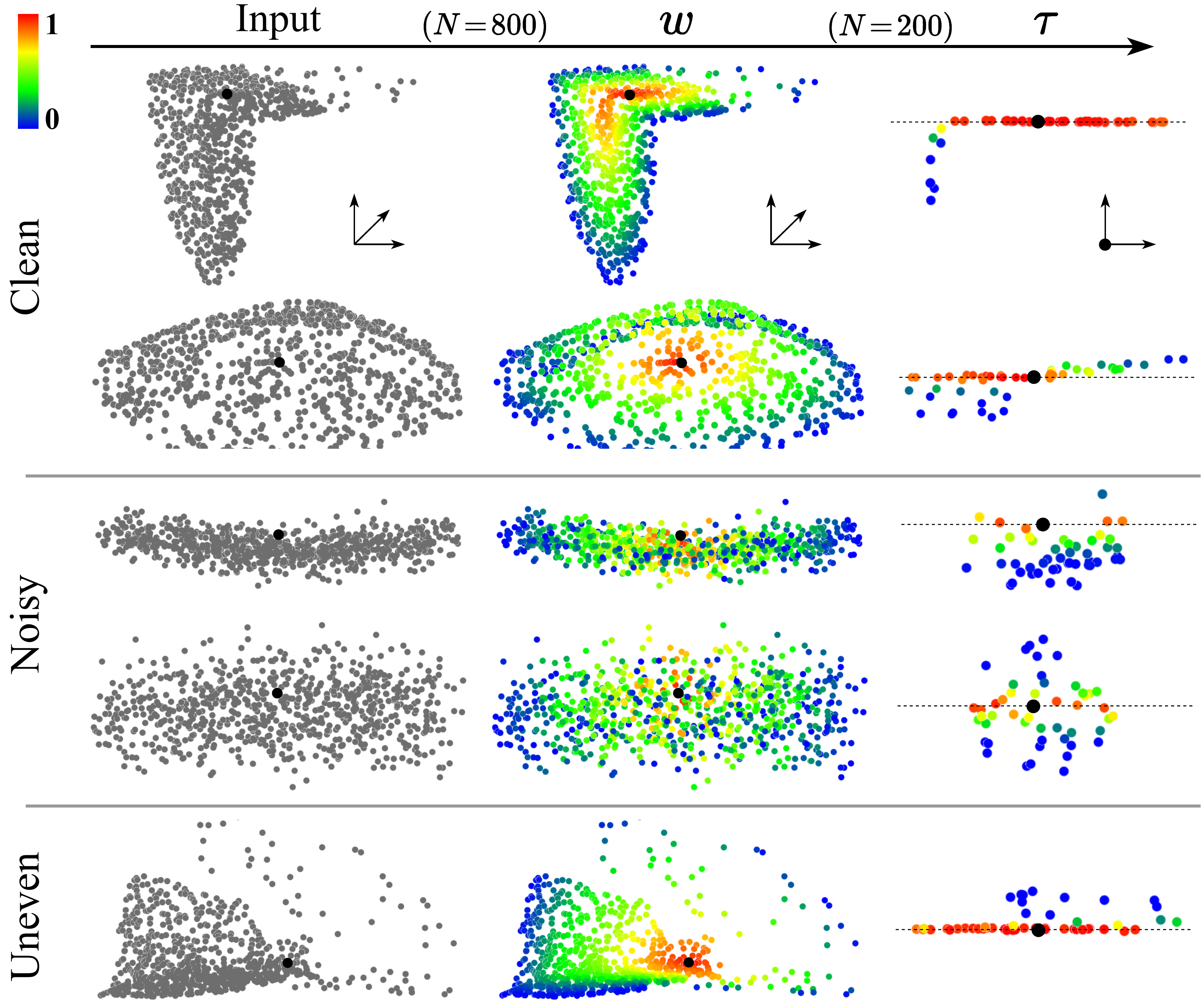}  \vspace{-0.5cm}
  \caption{
    Visualization of weights $w$ and $\tau$.
    It shows that the model's focus changes from central to coplanar with respect to the query point (black) at different stages.
    The perspective in the third column is changed for better visualization.
  }
  \label{fig:weights}
\end{figure}

\subsection{Complexity and Efficiency}
We compare the normal estimation methods that use deep neural networks.
As shown in Table~\ref{tab:time}, we report the number of the learnable network parameters of each method, and the inference time on the PCPNet dataset using NVIDIA 2080 Ti GPU.
Our method has the minimum network parameters and inference time.
Compared with the SOTA method, MSECNet~\cite{xiu2023msecnet}, our method has about $20\%$ of its network parameters and runs about $5.5$ times faster.

\subsection{Ablation Studies} \label{sec:ablation}

We provide the ablation results in Table~\ref{tab:ablation} (a)-(f), and each ablation experiment is discussed as follows.

\noindent
\textbf{(a) Per-point feature}.
The module in Eq.~\eqref{eq:unit} extracts per-point features from point-wise and local points through a two-branched structure.
If $\varphi$ is removed, it degenerates into a PointNet-like structure~\cite{qi2017pointnet}.
If $\phi$ is removed, it degenerates into a DGCNN-like structure~\cite{wang2019dynamic}.
The results show that the combination of these two structures is essential.

\noindent
\textbf{(b) Block $\mathcal{F}_1$}.
(1) We do not use the learnable distance-based weight $w$ in both $\mathcal{F}_1$ and $\mathcal{F}_2$.
(2) We adopt another weighting technique used in \cite{tombari2010unique}.
(3) We replace each layer $\mathcal{P}$ in $\mathcal{F}_1$ with MLP, \ie, without using $\mathcal{F}_1$.
The worse results of these ablations validate the effectiveness of our newly designed layer and show its key role in the normal estimation pipeline.

\noindent
\textbf{(c) Block $\mathcal{F}_2$}.
We replace each layer $\mathcal{P}$ in $\mathcal{F}_2$ with MLP, \ie, without using $\mathcal{F}_2$, or replace $\mathcal{F}_2$ with $\mathcal{F}_1$.
The results demonstrate that this block is important for improving the algorithm's performance.

\noindent
\textbf{(d) Cross-scale compensation}.
(1) We use \emph{softmax} in this module to generate attention weights for modulating individual feature channels (\emph{softmax}-1) or point channels (\emph{softmax}-2).
(2) We provide the results by replacing the entire cross-scale compensation module with the simple operation of concatenation or addition of features.
Different from the previous one, we also examine that we only replace $\eta(\cdot,\cdot)$ in Eq.~\eqref{eq:atten} from concat to add.
As we can see from the results, the performance of the algorithm decreases if \emph{softmax} is used.
The simple feature concatenation and addition operations or replacing $\eta(\cdot,\cdot)$ are inadequate for realizing the cross-scale feature compensation.

\noindent
\textbf{(e) Loss}.
We do not calculate the distances $d_{\rm sin}$ or $d_{\rm euc}$ of $\mathcal{L}_{\mathbf{n}}$ in Eq.~\eqref{eq:loss_n}. We also alternately leave the losses $\mathcal{L}_{\mathbf{n}}^p$ and $\mathcal{L}_{\tau}$ in Eq.\eqref{eq:finalloss} unused.
These ablations all lead to worse results.

\noindent
\textbf{(f) Input patch size}.
We train full models with a series of patch sizes $N\!=\!500,600,700,900,1000$.
A smaller size slightly degrades performance.
A larger size brings no performance improvement but consumes more time and memory.
To analyze the impact of input neighborhood size on algorithm performance, we provide more quantitative comparison results.
In Table~\ref{table:pcpnet_1}, the baseline methods are trained and tested with their default values, while our method is trained and tested with different patch sizes.
In Table~\ref{table:pcpnet_2}, all methods are trained with their default values but tested with different patch sizes.
Our method achieves excellent results even when training or testing with different patch sizes, while the baseline methods perform worse than our method under various input neighborhood sizes.

\noindent
\textbf{What does the model focus on?}
As shown in Fig.~\ref{fig:weights}, we visualize the learned weights $w$ in Eq.~\eqref{eq:weight} and $\tau$ in Eq.~\eqref{eq:output}.
We can see that the focus of our model changes along the normal estimation pipeline.
The model first focuses on points closer to the center during the feature aggregation, where features from large scales are transferred to small scales.
Then, the model focuses on some neighboring points coplanar with the query during the final normal prediction.

\section{Conclusion}

In this work, we analyze the effect of patch size on point cloud normal estimation, and propose a strategy that exploits the idea of patch feature fitting to approximate optimal features for normal estimation.
We use multi-scale features from different patch sizes to build the feature-based polynomial, and apply cross-scale attention to compensate for the approximation error.
The approximation strategy is implemented using an effective neural network, which aggregates features from multiple scales and achieves scale adaptation for varying local patches, to facilitate the geometric description around a point.
We conduct thorough experiments to compare with baselines and validate the proposed modules.
The main limitation of our method is that each point in the patch still needs a fixed neighborhood size to find its neighboring points to extract local features, which is also relatively time-consuming.
Future work includes exploring more efficient feature extraction techniques and more application scenarios of our method.

\section*{Acknowledgements}
This work was supported by the National Natural Science Foundation of China (62402401), the Sichuan Provincial Natural Science Foundation of China (2025ZNSFSC1462) and the Fundamental Research Funds for the Central Universities (2682025CX109).

\bibliographystyle{IEEEtran}
\bibliography{egbib}

\end{document}